\documentclass[sigplan,screen]{acmart}

\usepackage{xspace}
\usepackage[absolute,overlay]{textpos} 

\usepackage{microtype}
\usepackage{graphicx}
\usepackage{subcaption}
\usepackage{booktabs} 
\usepackage{multirow}
\usepackage[inline]{enumitem}
\usepackage{placeins} 
\usepackage{wrapfig}

\usepackage{hyperref}
\usepackage{xspace}
\usepackage{tabularx}
\usepackage{tikz}
\usetikzlibrary{positioning}

\usepackage{subcaption}
\usepackage{verbatim}
\usepackage{multirow}
\usepackage[makeroom]{cancel}
\usepackage{dsfont}
\usepackage{amsmath}
\usepackage{bbold}

\usepackage{tikz}  

\usepackage{tabularx}
\usepackage{color, colortbl}
\usepackage{hhline}

\usepackage[most]{tcolorbox}

\tcbset{
  insightbox/.style={
    colback=gray!10,    
    colframe=gray!50,   
    arc=2mm,            
    boxrule=0.8pt,      
    left=4pt, right=4pt, top=4pt, bottom=4pt, 
  }
}

\definecolor{commentgreen}{rgb}{0, 0.5, 0}
\usepackage[
  vlined,
  linesnumbered,
  ruled,
]{algorithm2e}
\DontPrintSemicolon{}
\SetKwComment{Comment}{$\triangleright$\ }{}

\SetCommentSty{mycommfont}

\let\oldnl\nl
\newcommand{\nonl}{\renewcommand{\nl}{\let\nl\oldnl}}

\usepackage{minted}

\usepackage{xcolor}

\usepackage{hyperref}
\hypersetup{
  colorlinks=true,      
  linkcolor=blue,       
  citecolor=magenta,    
  filecolor=cyan,       
  urlcolor=red          
}

\graphicspath{{figures/}}
\newenvironment{denseitemize}{
	\begin{itemize}[topsep=2pt, partopsep=0pt, leftmargin=1.5em]
		\setlength{\itemsep}{2pt}
		\setlength{\parskip}{0pt}
		\setlength{\parsep}{0pt}
	}{\end{itemize}}

\newcommand{\name}{TetriServe\xspace}

\AtBeginDocument{%
  }

\copyrightyear{2026}
\acmYear{2026}
\setcopyright{cc}
\setcctype{by}
\acmConference[ASPLOS '26]{Proceedings of the 31st ACM International Conference on Architectural Support for Programming Languages and Operating Systems, Volume 2}{March 22--26, 2026}{Pittsburgh, PA, USA}
\acmBooktitle{Proceedings of the 31st ACM International Conference on Architectural Support for Programming Languages and Operating Systems, Volume 2 (ASPLOS '26), March 22--26, 2026, Pittsburgh, PA, USA}
\acmDOI{10.1145/3779212.3790233}
\acmISBN{979-8-4007-2359-9/2026/03}

\begin{document}

\title{\name : Efficiently Serving Mixed DiT Workloads}

\author{Runyu Lu}
\authornote{Both authors contributed equally to this research.}
\email{runyulu@umich.edu}
\affiliation{%
  \institution{University of Michigan}
  \city{Ann Arbor}
  \state{Michigan}
  \country{USA}
}

\author{Shiqi He}
\authornotemark[1]
\email{shiqihe@umich.edu}
\affiliation{%
  \institution{University of Michigan}
  \city{Ann Arbor}
  \state{Michigan}
  \country{USA}
}

\author{Wenxuan Tan}
\email{wtan45@wisc.edu}
\affiliation{%
  \institution{University of Wisconsin-Madison}
  \city{Madison}
  \state{Wisconsin}
  \country{USA}
}

\author{Shenggui Li}
\email{shenggui001@e.ntu.edu.sg}
\affiliation{%
  \institution{Nanyang Technological University}
  \city{Singapore}
  \country{Singapore}
}

\author{Ruofan Wu}
\email{ruofanw@umich.edu}
\affiliation{%
  \institution{University of Michigan}
  \city{Ann Arbor}
  \state{Michigan}
  \country{USA}
}

\author{Jeff J. Ma}
\email{jeffjma@umich.edu}
\affiliation{%
  \institution{University of Michigan}
  \city{Ann Arbor}
  \state{Michigan}
  \country{USA}
}

\author{Ang Chen}
\email{chenang@umich.edu}
\affiliation{%
  \institution{University of Michigan}
  \city{Ann Arbor}
  \state{Michigan}
  \country{USA}
}

\author{Mosharaf Chowdhury}
\email{mosharaf@umich.edu}
\affiliation{%
  \institution{University of Michigan}
  \city{Ann Arbor}
  \state{Michigan}
  \country{USA}
}

\renewcommand{\shortauthors}{Runyu Lu et al.}

\begin{abstract}
Diffusion Transformer (DiT) models excel at generating high-quality images through iterative denoising steps, but serving them under strict Service Level Objectives (SLOs) is challenging due to their high computational cost, particularly at larger resolutions. 
Existing serving systems use fixed-degree sequence parallelism, which is inefficient for heterogeneous workloads with mixed resolutions and deadlines, leading to poor GPU utilization and low SLO attainment.

In this paper, we propose step-level sequence parallelism to dynamically adjust the degree of parallelism of individual requests according to their deadlines. We present \name \footnote{TetriServe is available at \url{https://github.com/DiT-Serving/TetriServe}.}, a DiT serving system that implements this strategy for highly efficient image generation. 
Specifically, \name introduces a novel round-based scheduling mechanism that improves SLO attainment by
(1) discretizing time into fixed rounds to make deadline-aware scheduling tractable, (2) adapting parallelism at the step level and minimizing GPU hour consumption, and (3) jointly packing requests to minimize late completions.
Extensive evaluation on state-of-the-art DiT models shows that \name achieves up to 32\% higher SLO attainment compared to existing solutions without degrading image quality. 

\end{abstract}



%


\maketitle

\section{Introduction}

\begin{figure}[t]
  \centering
    \includegraphics[width=0.35\textwidth]{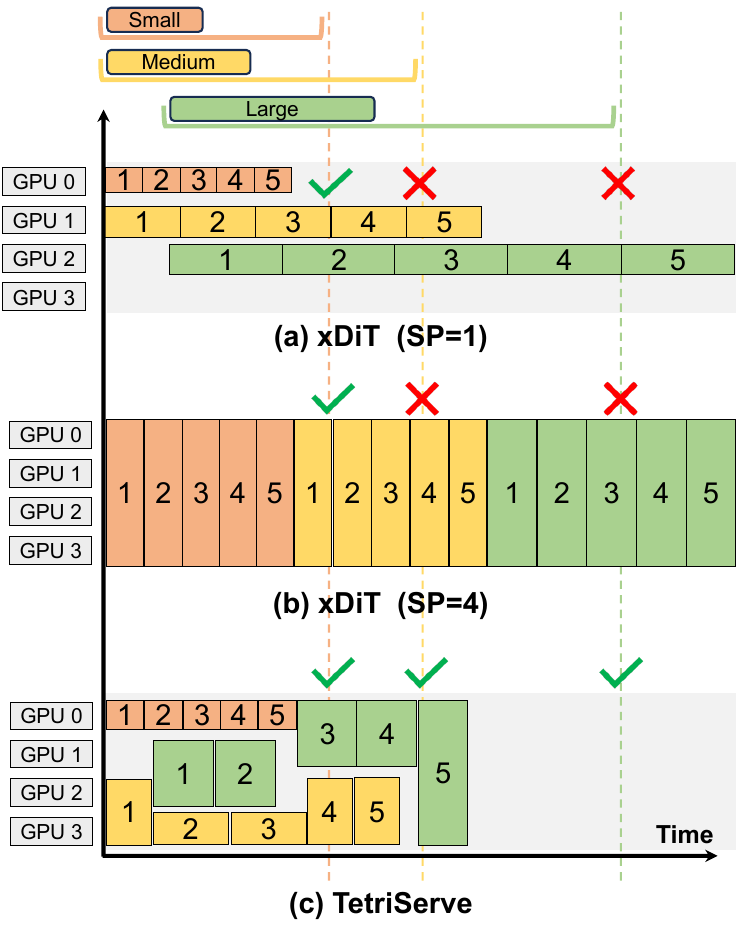}
    \vspace{-0.5em}
    \caption{Three DiT serving requests---each with 5 denoising steps---arrive over time with different SLOs and output resolutions. 
      DiT serving solutions using static parallelism cannot adapt and fail to meet multiple SLOs.
      \name meets more SLOs via SLO-aware scheduling and packing. 
    }
    \label{fig:toy_example}
    \vspace{-0.5em}
\end{figure}

Diffusion models \cite{ho2020denoising, sohl2015deep, song2021scorebased, DiT-iccv23, flux1dev2024, sd3, sd3.5} have significantly advanced text-to-image and text-to-video generation, enabling photorealistic content from natural language descriptions. 
They now power a wide range of commercial and creative services such as OpenAI Sora \cite{videoworldsimulators2024} and Adobe Firefly \cite{ApproximateCaching-nsdi24}. 
At the core of these breakthroughs are \emph{Diffusion Transformers (DiTs)} \cite{DiT-iccv23}, which have become the backbone of leading models including Stable Diffusion 3 (SD3) \cite{sd3} and FLUX.1-dev \cite{flux1dev2024}. 
By replacing conventional UNet architectures \cite{ronneberger2015unet, ho2020denoising}, DiTs achieve higher fidelity by iteratively refining a full-image latent representation over a sequence of discrete denoising steps, setting a new standard for generation quality.

As DiT models move into production, \emph{online DiT serving} becomes a key systems challenge. 
Deployments such as Flux AI \cite{flux1ai} must satisfy strict service level objectives (SLOs) in the form of a \emph{deadline} for each request while sharing a fixed GPU pool across many users to minimize cost. 
Serving is particularly challenging because requests arrive with heterogeneous output resolutions and tight deadlines. 

Despite advances in LLM serving \cite{loongserve-sosp24, PagedAttention-sosp23, SGLang-neurips24, prefillonly-sosp25, splitwise-isca24, SpecInfer-asplos24, Miao2024SpotServe, Mei2025Helix}, these solutions are insufficient: DiTs have fundamentally different serving characteristics.
Specifically, DiT inference differs from LLMs in three ways: (i) it is stateless, requiring no KV cache; (ii) it is compute-bound, as multiple denoising steps operate on the full set of latent image tokens; and (iii) model sizes are small enough to fit on a single GPU.
Consequently, generating a high-resolution $2048\times2048$ image on a single H100 GPU can take up to a minute, while a $4096\times4096$ image may exceed ten minutes.
To meet the stringent latency demands of online serving, parallelism is essential.

The most common approach for parallelizing DiTs is \emph{sequence parallelism (SP)} \cite{DeepSpeed-Ulysses-arxiv23, Ring-Attention-arxiv23}, which partitions the sequence of image tokens across GPUs.
However, simply applying a fixed degree of SP to all requests is inefficient and leads to poor SLO attainment.
This is because the optimal degree of parallelism is highly sensitive to the input image resolution; a configuration that is ideal for one resolution can be detrimental to another.
As shown in the toy example in Figure~\ref{fig:toy_example}, the fixed-degree SP approach creates a fundamental tradeoff: low degrees of parallelism (e.g., SP=1 or 2) are efficient for small inputs but underutilize the GPU cluster for large ones by leaving some GPUs idle and prolonging request runtime, while high degrees of parallelism (e.g., SP=4 or 8) accelerate large inputs but introduce excessive communication overhead for small ones, leading to head-of-line blocking.
Compounding this issue, existing DiT inference engines \cite{xDiT-arxiv24} are non-preemptive: once a request begins execution with a fixed degree of parallelism, it holds its allocated GPU(s) until completion, preventing more optimal scheduling of other requests in the queue.

We observe that \emph{step-level scheduling}, in which the degree of parallelism is adjusted across steps within each request based on its resolution and deadline, can significantly improve the serving efficiency of mixed DiT workloads. 
High-resolution or urgent requests can be accelerated with more GPUs, while smaller or less urgent ones conserve resources. 
Unfortunately, we prove that finding a globally optimal step-level schedule that maximizes deadline satisfaction under a fixed GPU budget is NP-hard (\S\ref{sec:problem}). 
In addition, the online arrival of requests and the need for millisecond-level scheduling decisions make exhaustive optimization infeasible.


We present \emph{\name}, a step-level DiT serving system designed to maximize SLO attainment under deadline constraints. 
At its core, \name introduces a \emph{deadline-aware round-based scheduler} that transforms the continuous time in the serving problem into a sequence of tractable, fixed-duration rounds. 
In each round, the scheduler decides which requests to serve and at what GPU parallelism degree. 
To make these decisions, \name leverages a cost model that profiles per-step latency as a function of GPU count and identifies the \emph{minimal feasible GPU allocation} for each request that can still meet its deadline. 
This allows \name to construct a set of candidate allocations and perform request packing with the explicit goal of minimizing the number of requests that would otherwise become late in the next round.


\name further enhances GPU efficiency while preserving request deadlines. 
It uses \emph{selective continuous batching} to merge steps across small-resolution requests, reducing kernel launch overhead and boosting throughput. 
Meanwhile, \emph{GPU placement preservation} and \emph{work-conserving elastic scale-up} ensure idle GPUs are utilized without remapping distributed jobs. 
Together with the round-based scheduler, these techniques allow \name to handle diverse DiT workloads---from small to large resolutions---while substantially improving deadline satisfaction over fixed-degree baselines.

We evaluate \name on popular open-source DiT models (FLUX.1-dev and SD3) and different hardware platforms ($8\times$H100 and $4\times$A40 nodes). 
We show that \name consistently outperforms xDiT \cite{xDiT-arxiv24}---a DiT-serving engine that allows different fixed SP configurations---across diverse experimental settings by up to 32\% in terms of SLO attainment ratio. 
\name is also robust to bursty request arrival patterns, diverse workload mixes, and different model--hardware combinations.

We summarize the contributions as follows:
\begin{itemize}
  \item We cast DiT serving as a step-level GPU scheduling problem and prove its NP-hardness. 
  \item We present \name, a deadline-aware round-based scheduler that minimizes late completions via dynamic programming.
  \item We show that \name achieves substantial gains in SLO attainment over fixed-degree baselines on state-of-the-art DiT models while maintaining image quality. 
\end{itemize}

\section{Motivation}
\label{sec:motivation}
Serving DiT models has become a popular workload for modern image generation systems \cite{ApproximateCaching-nsdi24, xDiT-arxiv24}. 
DiT inference is both compute-intensive and latency-sensitive. 
To better understand the challenges of serving such workloads, in this section, we discuss DiT background, workload characteristics, and the resulting opportunities and challenges.
 

\subsection{DiT Background}
\label{sec:dit-background}


Diffusion models \cite{ho2020denoising, sohl2015deep, song2021scorebased, DiT-iccv23, videoworldsimulators2024} have significantly advanced text-to-image and text-to-video generation, enabling photorealistic content from natural language descriptions.
Each step operates on the full latent representation, removing noise based on a learned denoising function.
Although early diffusion models used \emph{UNet} architectures \cite{ronneberger2015unet, ho2020denoising}, modern high-quality image generators use \emph{Diffusion Transformers (DiTs)} \cite{DiT-iccv23, dosovitskiy2021vit} as their backbone.
DiTs use attention \cite{vaswani2017attention} to capture global context and long-range dependencies.

\paragraph{DiT vs.\ LLM Parallelism} 
Although both DiTs and LLMs are built upon the Transformer architecture, their inference characteristics diverge significantly, requiring different parallelism strategies.
Traditional model-sharding strategies for LLMs, such as tensor and pipeline parallelism, are inefficient for DiTs. This is because DiT models are typically small enough to fit on a single GPU. For example, the largest open-source text-to-image DiT has only 12B parameters \cite{flux1dev2024} and fits comfortably on a single 80GB H100 GPU. Consequently, applying model sharding introduces unnecessary communication overhead without the benefit of accommodating a larger model, resulting in poor hardware utilization.

DiTs adopt \emph{sequence parallelism (SP)} \cite{DeepSpeed-Ulysses-arxiv23, Ring-Attention-arxiv23, li2022sequenceparallelismlongsequence}, a more efficient parallel approach tailored to their compute-bound nature. 
In SP, token sequences (image tokens) are distributed across GPUs, enabling collaborative computation within each transformer layer. 
Two representative implementations are \emph{Ulysses attention} \cite{DeepSpeed-Ulysses-arxiv23}, which uses all-to-all collectives to transpose tokens and heads across GPUs before local attention, and \emph{Ring attention} \cite{Ring-Attention-arxiv23}, which arranges GPUs in a ring and passes partial Q, K, V slices peer-to-peer, overlapping communication with computation. In practice, Ulysses attention is often preferred on systems with high-bandwidth interconnects like NVLink, as its use of collective primitives can be more efficient \cite{xDiT-arxiv24}.

\subsection{Characteristics of DiT Workloads}
\label{sec:dit-characteristics}

DiT serving exhibits distinctive workload characteristics that affect the design of scheduling and resource management. 

\paragraph{Heterogeneous Inputs.}
Unlike LLM workloads, where input text can vary widely in length, DiT serving workloads are characterized by a small, discrete set of possible input image resolutions~\cite{flux1ai, stabilityai-api}. In this work, we focus on four representative resolutions common in production environments; their characteristics for the FLUX.1-dev model \cite{flux1dev2024} are detailed in Table~\ref{tab:flux-input-sizes}. Despite the small number of distinct input sizes, the substantial differences in their computational demands still lead to highly heterogeneous resource requirements across requests.

\begin{table}
    \centering
    \footnotesize
    \caption{Characteristics of representative input sizes for the FLUX.1-dev model \cite{flux1dev2024}, including latent tokens and computational cost (TFLOPS). Execution stability (CV) is measured over 20 steps on 8xH100 GPU for different sequence parallelism (SP) degrees.}
    \label{tab:flux-input-sizes}
    \begin{tabular}{crrrrrr}
        \toprule
        \textbf{Image Size} & \textbf{Tokens} & \textbf{TFLOPs} & \textbf{SP=1} & \textbf{SP=2} & \textbf{SP=4} & \textbf{SP=8} \\
        \midrule
        $256 \times 256$    & 256    &   556.48  & 0.13\%  & 0.31\% & 0.67\% & 0.62\% \\
        $512 \times 512$    & 1024    &  1388.24  & 0.06\%  & 0.15\% & 0.14\% & 0.53\% \\
        $1024 \times 1024$  & 4096   &  5045.92  & 0.07\%  & 0.12\% & 0.04\% & 0.09\% \\
        $2048 \times 2048$  & 16384   & 24964.72  & 0.05\% & 0.11\% & 0.14\% & 0.28\% \\
        \bottomrule
    \end{tabular}
\end{table}
\FloatBarrier

\paragraph{Predictable Execution.}
Despite input diversity, DiT inference remains compute-bound and therefore exhibits stable per-step runtimes across a wide range of input resolutions. 
As shown in Table~\ref{tab:flux-input-sizes}, execution time is highly stable: profiling over 100 runs with varying sequence-parallel degrees yields a coefficient of variation (CV) below 0.7\% in all cases. 
This low variability indicates that DiT model inference is predictable across resolutions and degrees of parallelism, enabling accurate performance modeling and effective deadline-aware scheduling.

\begin{tcolorbox}[insightbox]
\textbf{Insight 1:} \emph{DiT workloads consist of heterogeneous input requests with  different output resolutions, but per-step runtime for each resolution is highly predictable.}
\end{tcolorbox}


\begin{figure}[!t]
    \centering
    \includegraphics[width=0.9\columnwidth]{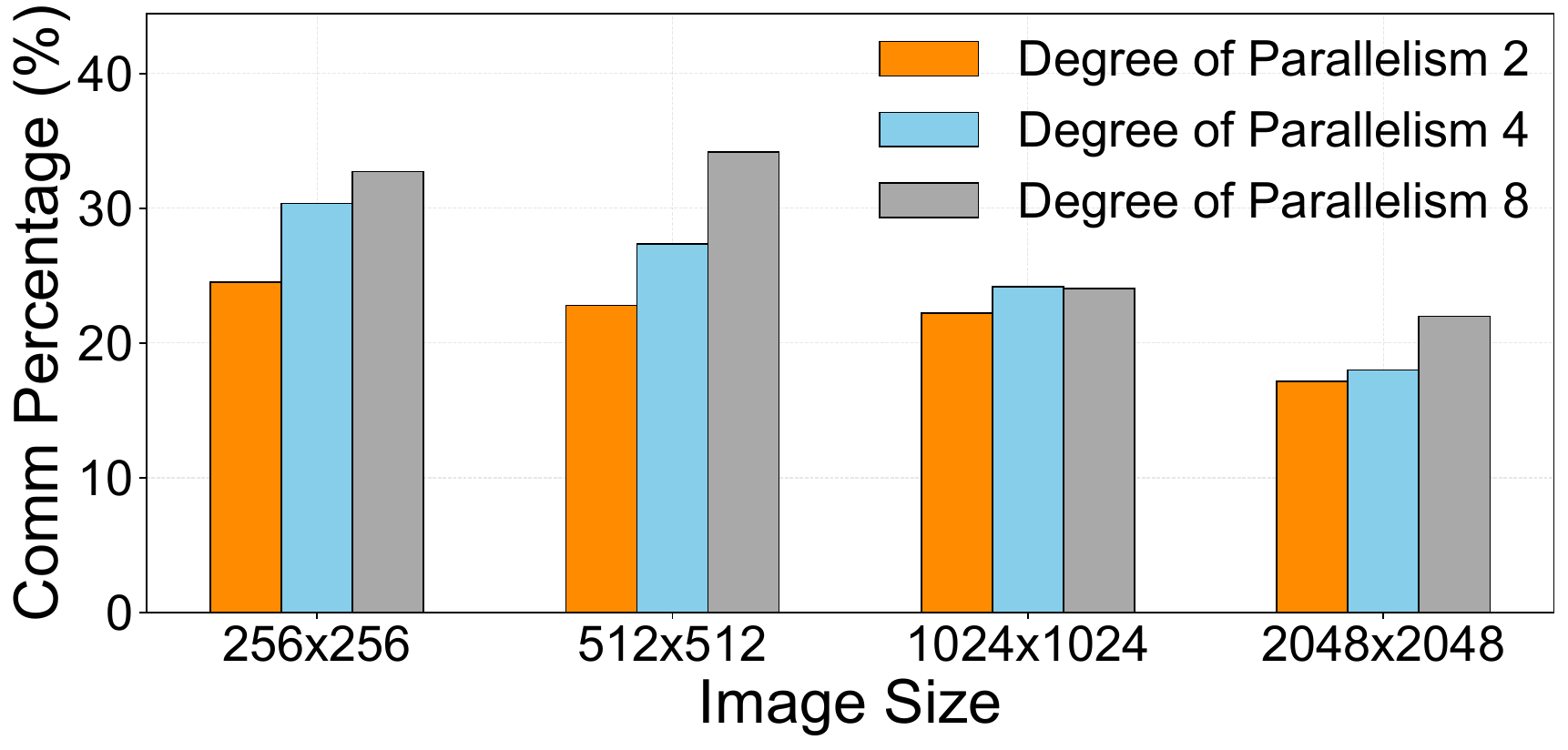}
    \caption{Percentage of time spent in communication for FLUX.1-dev for four resolutions on an $8\times$H100 server (Batch Size = 4). Larger resolutions benefit more from increased parallelism because of relatively less communication overhead.}
    \label{fig:comm_comp_ratio}
\end{figure}

\begin{figure}[t]
    \centering
    \includegraphics[width=0.7\columnwidth]{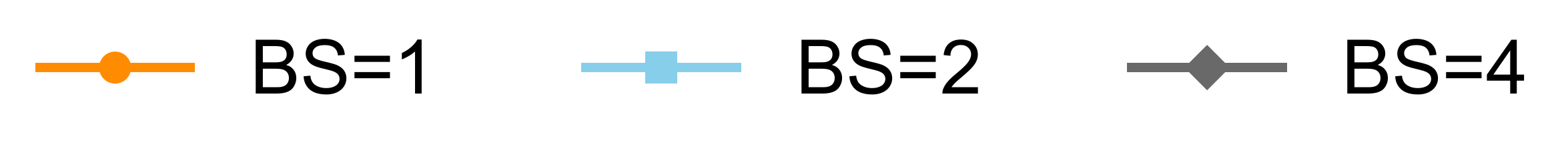}
    \vspace{-1em}
    \begin{subfigure}[b]{0.45\columnwidth}
        \centering
        \includegraphics[width=\linewidth]{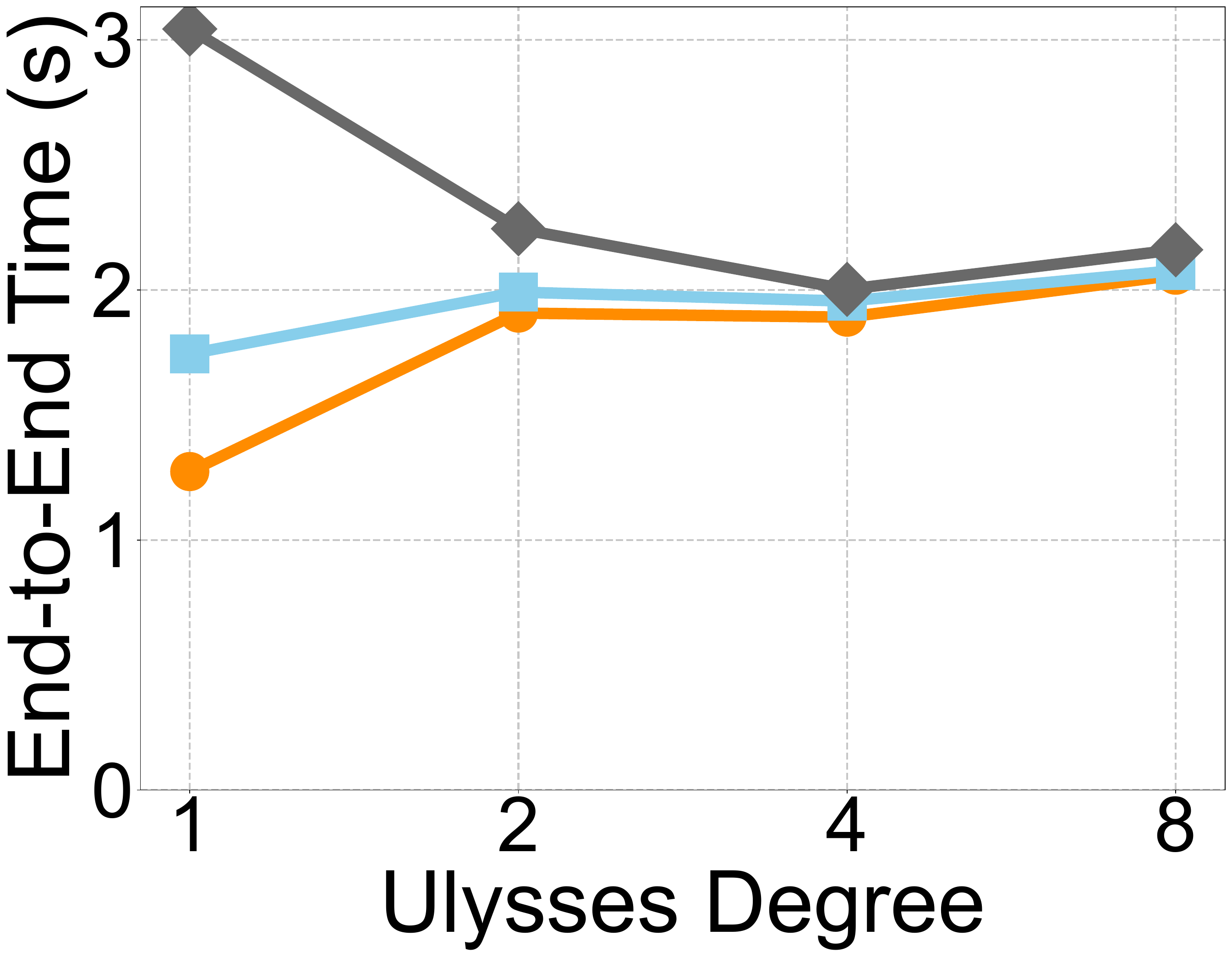}
        \caption{Image Size: 256$\times$256}
        \label{fig:efficiency_256}
    \end{subfigure}
    \hfill
    \begin{subfigure}[b]{0.45\columnwidth}
        \centering
        \includegraphics[width=\linewidth]{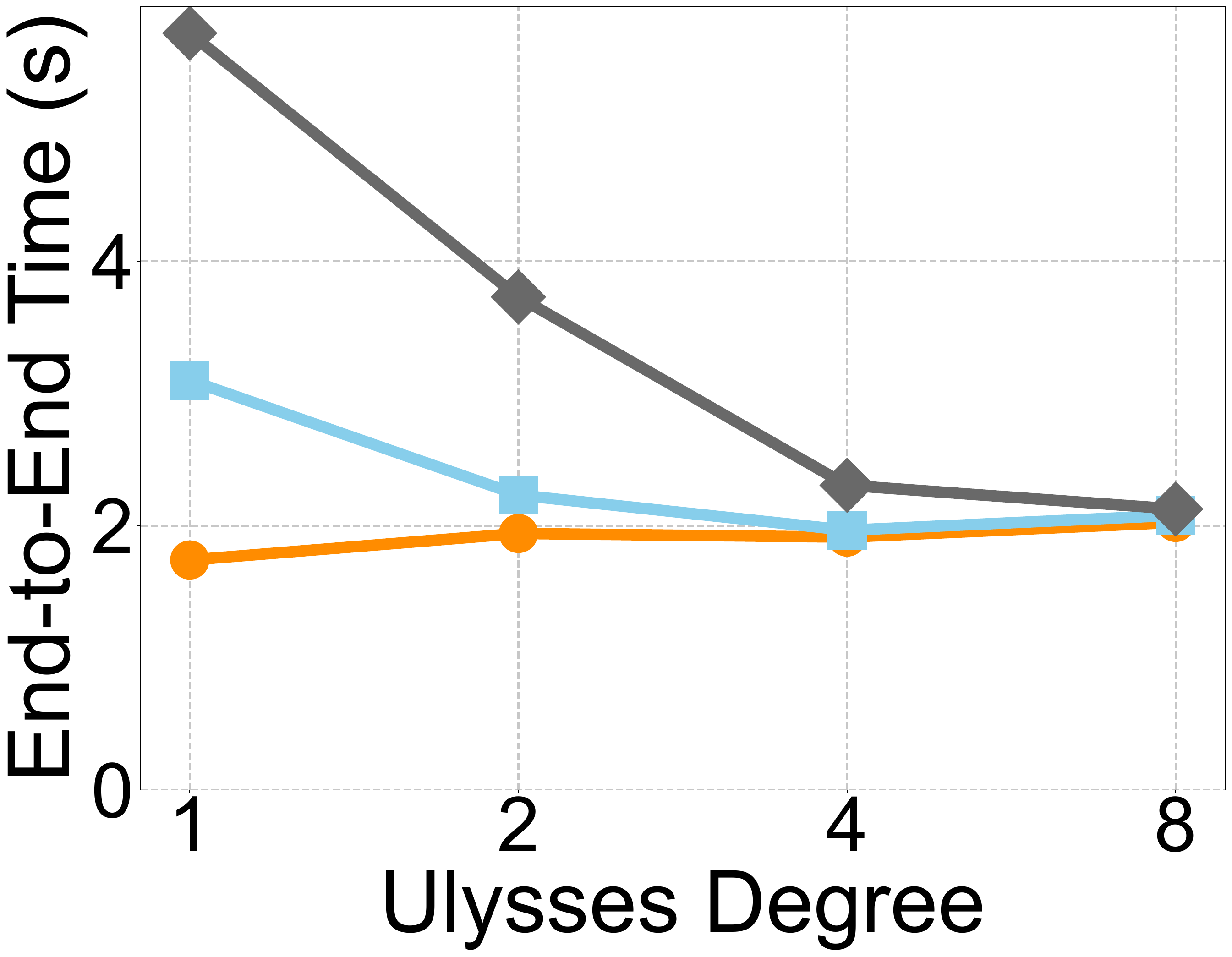}
        \caption{Image Size: 512$\times$512}
        \label{fig:efficiency_512}
    \end{subfigure}
    
    \vspace{2em}
    
    \begin{subfigure}[b]{0.45\columnwidth}
        \centering
        \includegraphics[width=\linewidth]{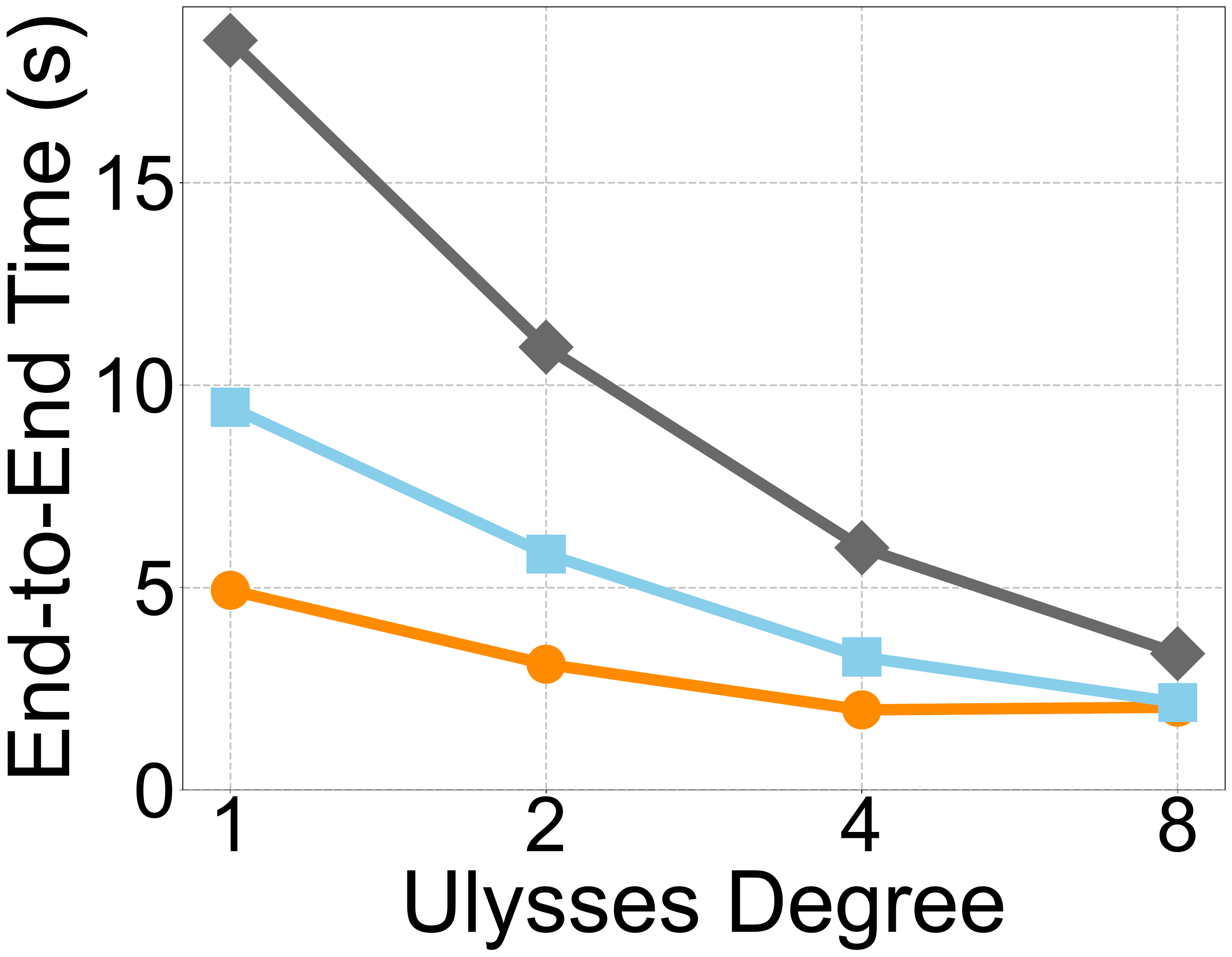}
        \caption{Image Size: 1024$\times$1024}
        \label{fig:efficiency_1024}
    \end{subfigure}
    \hfill
    \begin{subfigure}[b]{0.45\columnwidth}
        \centering
        \includegraphics[width=\linewidth]{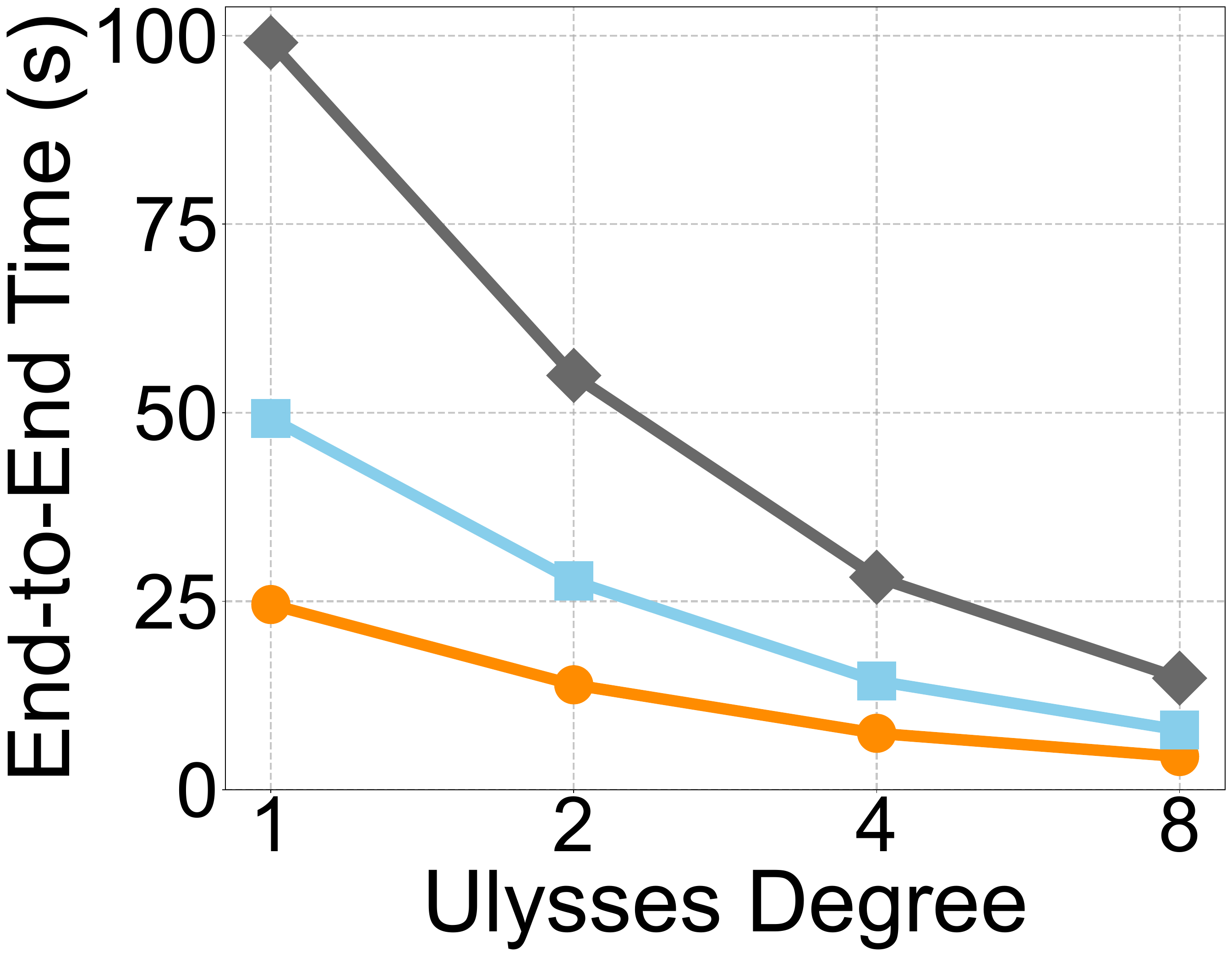}
        \caption{Image Size: 2048$\times$2048}
        \label{fig:efficiency_2048}
    \end{subfigure}

    \caption{End-to-end scaling efficiency of FLUX.1-dev for four resolutions on an $8\times$H100 server for different batch size (BS). 
        Efficiency scales sublinearly. 
        Larger resolutions benefit more from increased parallelism, while smaller resolutions exhibit limited scalability.
        Note different Y-axes scales.}
    \label{fig:e2e_scaling_efficiency}
\end{figure}


\paragraph{Scaling Efficiency of Sequence Parallelism.}
Sequence parallelism distributes tokens across GPUs, but its scaling efficiency is sublinear to the degree of parallelism. 
Two factors drive this: (i) communication overhead from collectives (all-to-all or ring exchanges) that scales with the degree of parallelism and sequence length; and 
(ii) reduced per-GPU kernel efficiency when workloads are split, lowering occupancy and cache locality. 
Figure~\ref{fig:comm_comp_ratio} quantifies this by showing the communication percentage across image sizes and degrees of parallelism. 
For small inputs (e.g., $256 \times 256$ and $512 \times 512$), increasing the degree of parallelism rapidly increases the communication percentage, exceeding 30\% at higher degrees. In this case, communication dominates execution time, leading to poor scaling and decreasing the benefits from additional GPUs.
Figure~\ref{fig:e2e_scaling_efficiency} shows that small inputs (e.g., $256 \times 256$, $512 \times 512$) underutilize GPUs and scale poorly, while larger inputs (e.g., $1024 \times 1024$, $2048 \times 2048$) improve efficiency though computation remains the bottleneck. 
This explains why in Figure~\ref{fig:toy_example}, latency does not scale linearly with the number of GPUs.

\begin{tcolorbox}[insightbox]
\textbf{Insight 2:} \emph{Sequence parallelism in DiT workloads scales sublinearly with the degree of parallelism and differently for each input resolution.}
\end{tcolorbox}



\subsection{Challenges and Opportunities}
\label{sec:opportunities}

\paragraph{Limitations of Current Solutions.}

\begin{figure}[t]
    \centering
    \includegraphics[width=0.98\columnwidth]{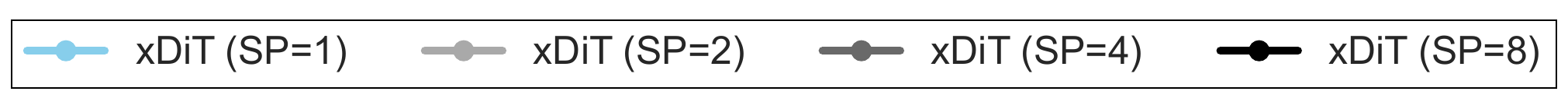}
    \vspace{-1em}
    \begin{subfigure}[b]{0.48\columnwidth}
        \centering
        \includegraphics[width=\linewidth]{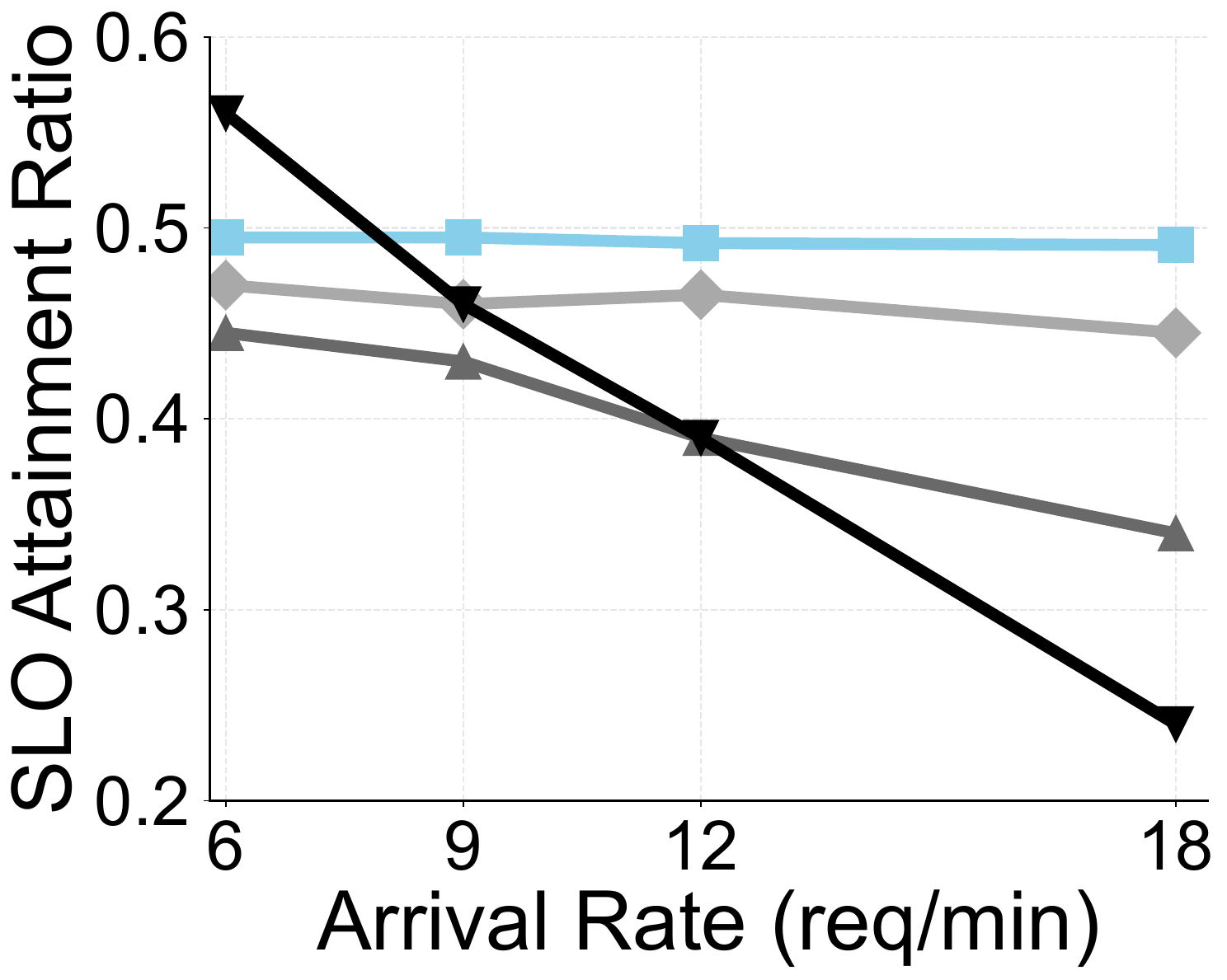}
        \caption{Fixed Degree Performance}
        \label{fig:even_fixed_row:ratio}
    \end{subfigure}
    \hfill
    \begin{subfigure}[b]{0.48\columnwidth}
        \centering
        \includegraphics[width=\linewidth]{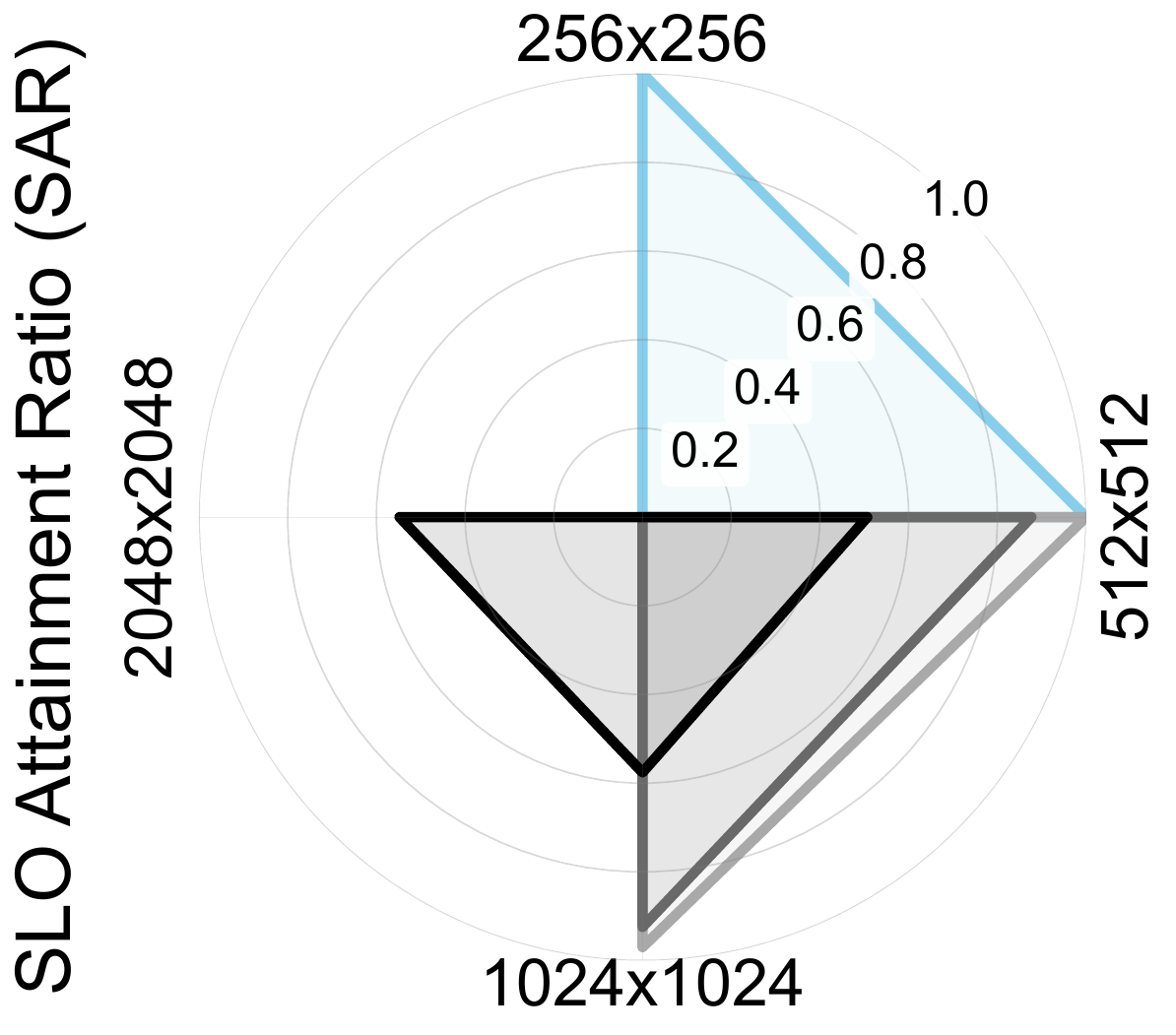}
        \caption{Breakdown by Resolution}
        \label{fig:even_fixed_row:spider}
    \end{subfigure}
    \caption{Performance of fixed degree xDiT variants under the Uniform workload. (a) The overall SLO Attainment Ratio (SAR) is low for all fixed strategies. (b) The spider plot, shown for a representative arrival rate of 12 req/min, reveals the underlying reason: low SP degrees fail on large resolutions, while high SP degrees perform poorly on small ones.}
    \label{fig:even_fixed_row}
\end{figure}

Conventional serving strategies using a fixed degree of parallelism are ill-suited for the heterogeneous nature of DiT workloads, a limitation illustrated in the toy example in Figure~\ref{fig:toy_example}. 
With data parallelism (xDiT, SP=1), the small request meets its deadline, but the larger requests fail due to insufficient processing speed. 
Conversely, a high fixed degree of parallelism (xDiT, SP=4) handles the large request well, but it still misses the deadline (along with the medium one) due to head-of-line blocking and inefficient resource use of the small request.

Experimental results confirm this trade-off. 
As shown in Figure~\ref{fig:even_fixed_row:ratio}, under a Uniform workload with a tight SLO Scale of $1.0\times$, no fixed-parallelism strategy achieves an SLO Attainment Ratio (SAR) above 0.6. 
The spider plot in Figure~\ref{fig:even_fixed_row:spider} reveals why: each fixed strategy only works well for specific resolutions. 
\emph{SP=1} and \emph{SP=2} achieve near-perfect SAR for $256\times256$ images but fail completely for $2048\times2048$, while \emph{SP=4} and \emph{SP=8} handle $2048\times2048$ effectively but perform poorly on smaller resolutions due to scaling inefficiency and head-of-line blocking. 
No single parallelism degree works across the board.


\paragraph{Optimization Opportunities.}
The limitations of fixed parallelism highlight a key opportunity: moving to dynamic, \emph{step-level sequence parallelism}. As shown in Figure~\ref{fig:toy_example}(c), our approach, \name, meets all three deadlines by adapting the degree of parallelism for each request at the step level. It assigns fewer GPUs to the initial steps of the medium request, freeing up resources, and then scales up to meet the deadline, thus avoiding the rigid trade-offs of fixed strategies.

This flexibility to adjust the sequence parallelism degree \emph{per step} allows a scheduler to allocate more GPUs when deadlines are tight and fewer when they are not, freeing capacity for other requests. By exploiting DiTs' predictable step execution times and heterogeneous scaling behavior, this approach enables finer-grained resource shaping and better SLO attainment than conventional fixed-SP policies.

\begin{tcolorbox}[insightbox]
\textbf{Insight 3:} \emph{Step-level parallelism adapts GPU allocation to request deadlines, avoiding the resource waste of fixed parallelism and improving SLO attainment.}
\end{tcolorbox}

\section{\name Overview}

\label{sec:overview}

\name allows more DiT serving requests with heterogeneous output resolutions to meet their SLOs by judiciously scheduling and packing them on shared GPU resources.
In this section, we provide an overview of how \name fits in the DiT serving lifecycle to help the reader follow the subsequent sections.

\paragraph{System Components.} 
\name is designed around a scheduler that makes deadline-aware GPU allocation decisions in a round-based manner. 
Its key components are:

\begin{itemize}
    \item \textbf{Request Tracker:} Maintains metadata on active requests, including resolutions, deadlines, and execution states (e.g., remaining steps).
    
    \item \textbf{Scheduler:} The core component consists of \emph{deadline-aware GPU allocation} and \emph{round-based request packing}. 
        At every round, it minimizes individual requests' GPU consumption while maximizing SLO attainment.
    
    \item \textbf{Execution Engine:} A distributed pool of GPU workers that execute assigned diffusion steps in parallel.
    
    \item \textbf{Latent Manager:} Handles intermediate latent representations across steps, reducing redundant computation and memory overhead.
\end{itemize}

Together, these components enable \name to adapt resource allocation at millisecond scale, sustaining high throughput and SLO attainment for heterogeneous DiT workloads.

\begin{figure}[t]
    \centering
    \includegraphics[width=\linewidth]{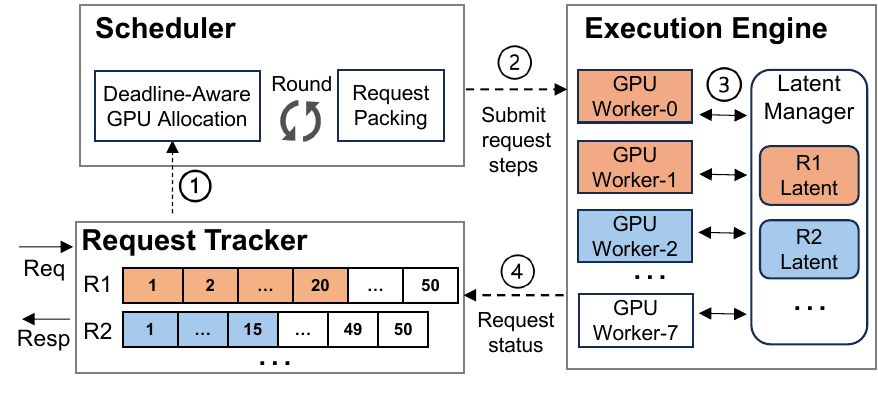}
    \caption{\name architecture and request lifecycle.}
    \label{fig:system-design}
\end{figure}

\paragraph{Request Lifecycle.}
When a request arrives, the \textit{Request Tracker} records its resolution, state, and deadline. The \textit{Scheduler} then places it into the next scheduling round \textcircled{1}, where a deadline-aware policy determines GPU allocations in terms of step numbers for each request for one round. 
For example, in Figure~\ref{fig:system-design}, it selects Request 1 to run 20 steps on 2 GPUs (orange) and Request 2 to run 15 steps on 1 GPU (blue) for the scheduling round. 
Different requests are dispatched to GPU workers in the \textit{Execution Engine} \textcircled{2}, which compute diffusion steps and produce intermediate latents managed by the \textit{Latent Manager} \textcircled{3}. Upon completion, workers notify the request tracker to update dependent steps \textcircled{4}. After all steps finish, the final output is returned to the user.

\section{Deadline-Aware Round-Based Scheduler}
\label{sec:design}
\name introduces a deadline-aware scheduler designed to optimize SLO attainment for DiT serving.
We begin with a formal definition of the GPU scheduling problem in the offline scenario and prove that it is NP-hard. 
We then propose a \textit{round-based scheduling mechanism}, which maximizes goodput via minimizing GPU-hour consumption for each request. 
Later we proposes enhancements so that \name balances utilization, latency, and scalability in DiT serving.


\subsection{Problem Statement}  
\label{sec:problem}
Given a collection of GPUs and requests, the DiT serving objective for each invocation of the scheduler is the following:  
\emph{Find a step-level schedule that maximizes the number of requests meeting their deadlines given a fixed number of GPUs.}


\paragraph{Problem Formulation}

Consider an \( N \)-GPU cluster and \( R \) outstanding requests.
Each request \( req_i \) consists of a sequence of \( S_i \) dependent diffusion steps \( \{ s_{i1}, s_{i2}, \ldots, s_{iS_i} \} \). 
Each step \( s_{ij} \) can be executed using \( k \in \{1, 2, 4, \ldots, N\} \) GPUs, where \( k \) is a power of two.
The execution time of a step, denoted \( T_{ij}(k) \), is a function of \( k \). 
The completion time of a request is defined as:
\[
C_i = \sum_{j=1}^{S_i} \big[ Q_{ij} + T_{ij}(A_{ij}) \big],
\]
where \( Q_{ij} \) is the queuing delay before step \( s_{ij} \) begins and \( A_{ij} \) is the number of GPUs allocated. Then we can formulate the DiT serving objective as:
\[
\text{Maximize} \quad \sum_{i=1}^{R} I_i, \quad \text{where } I_i = \begin{cases} 
1 & \text{if } C_i \leq D_i, \\
0 & \text{otherwise.}
\end{cases}
\]

This formulation is subject to the following conditions:
\begin{enumerate}
    \item \textbf{Step Dependency}: A step \( s_{ij} \) can start only after the previous step completes:
    \[
    \text{Start}(s_{ij}) \geq \text{Completion}(s_{i(j-1)}). \quad \forall i, \forall j > 1
    \]
    Therefore, at most one step of a request can be executed at any time.
    \item \textbf{GPU Capacity}: At any time, the total number of GPUs allocated across all steps cannot exceed \(N\):
    \[
    \sum_{i=1}^R \sum_{j=1}^{S_i} A_{ij}(t) \leq N, \quad \forall t
    \]
    where \(A_{ij}(t)\) denotes the GPUs allocated to step \( s_{ij} \) if it is running at time \(t\), and zero otherwise.
\end{enumerate}

The goal is to find a set of GPU assignments \( \{ A_{ij} \} \) that maximizes the number of requests meeting their deadlines.

\paragraph{NP-hardness.} To highlight the computational complexity, we consider the special case where each request has a single non-preemptive step (\(S_i=1\)). Time is discretized into slots \(\mathcal{T}=\{0,1,\ldots,T_{\max}-1\}\). Let \(\mathcal{K}=\{1,2,4,\ldots,N\}\) denote the allowed GPU allocations. For each request \(i\), start time \(t \in \mathcal{T}\), and GPU count \(k \in \mathcal{K}\), introduce a binary decision variable:
\[
x_{i,t,k} =
\begin{cases}
1 & \text{if request $i$ starts at time $t$ with $k$ GPUs}, \\
0 & \text{otherwise}.
\end{cases}
\]

\textbf{Objective.} Maximize the number of requests completing by their deadlines:
\[
\max \sum_{i} \sum_{t \in \mathcal{T}} \sum_{k \in \mathcal{K}} x_{i,t,k}.
\]

\textbf{Constraints.}
\begin{align}
\sum_{t \in \mathcal{T}} \sum_{k \in \mathcal{K}} x_{i,t,k} &\leq 1, && \forall i, \\
arrival\_time(i) & \leq t, && \forall i, \\
t + T_i(k) &\leq D_i, && \forall i,t,k, \\
\sum_{i} \sum_{k \in \mathcal{K}} \sum_{u \in [t,\,t+T_i(k)-1]}
k \cdot x_{i,t,k} &\leq N, && \forall t,u \in \mathcal{T}, \\
x_{i,t,k} &\in \{0,1\}, && \forall i,t,k.
\end{align}

Constraint (1) ensures each request starts at most once. Constraints (2) and (3) enforce arrival times and deadline feasibility. Constraint (4) enforces that at any time slot \(u\), the sum of GPUs assigned to running requests does not exceed system capacity \(N\). Constraint (5) enforces integrality. 

This Zero-one Integer Linear Program (ZILP) exactly captures the offline DiT serving problem in the single-step case, where \( I_i = \sum_{t \in \mathcal{T}} \sum_{k \in \mathcal{K}} x_{i,t,k}\). 
We show in Appendix~\ref{appendix:appendix-proofs} that solving such formulations is NP-hard~\cite{garey1977two, bar1999approximating, papadimitriou1998combinatorial,zhang2023shepherd,khare2025superserve}. Therefore, \textbf{multi-step DiT serving is NP-hard} as well.

\begin{table}[t]
\centering
\caption{Notations used in the GPU Scheduling Problem.}
\begin{tabular}{lp{0.35\textwidth}}
\toprule
\textbf{Symbol} & \textbf{Description} \\
\midrule
\( N \) & Total number of GPUs. \\
\( R \) & Number of requests. \\
\( S_i \) & Number of steps in request \( req_i \). \\
\( D_i \) & Deadline of request \( req_i \). \\
\( T_{ij}(k) \) & Execution time of step \( s_{ij} \) with \( k \) GPUs. \\
\( Q_{ij} \) & Queueing delay before step \( s_{ij} \) starts. \\
\( A_{ij} \) & GPU allocation for step \( s_{ij} \). \\
\( C_i \) & Completion time of request \( req_i \). \\
\bottomrule
\end{tabular}
\end{table}

\subsection{Round-Based Scheduling}
Step-level scheduling for DiT serving is NP-hard, making global optimization expensive. 
To enable practical scheduling, \name adopts a round-based heuristic: instead of scheduling steps arbitrarily in a continuous global timeline, \emph{we discretize execution into rounds}, where each round corresponds to a fixed-length GPU execution window. 
This allows us to \textit{(i) limit the scheduling search space} and \emph{(ii) enable efficient preemption between rounds}. 
Within each round, \name determines the minimal required GPU allocation for requests and dynamically packs these requests to maximize SLO attainment ratio. 

\subsubsection{Deadline-Aware GPU Allocation}




Exhaustively enumerating GPU allocations for each step is infeasible, and over-allocation wastes resources due to scaling inefficiencies in DiT models (e.g., kernel launch and communication overheads).  
While more GPUs reduce latency, they increase total GPU hours. 
To balance these trade-offs, \name identifies the minimal GPU allocation needed for each request to meet its deadline at the beginning of each round. 
Since required allocation depends mainly on resolution and deadline, this approach avoids exploring the full allocation space.

For a step \( s_{ij} \), the execution time \( T_{ij}(k) \) is a function of the number of GPUs \( k \).
The GPU hour for executing step \( s_{ij} \) with \( k \) GPUs is \(k \times T_{ij}(k)\). The goal is to minimize the total GPU hour for each request:
\[
\min_{\{A_{ij}\}} \; \sum_{j=1}^{S_i} \big( A_{ij} \times T_{ij}(A_{ij}) \big) 
\quad \text{s.t.} \quad 
\sum_{j=1}^{S_i} \big( Q_{ij} + T_{ij}(A_{ij}) \big) \leq D_i
\]
where \(A_{ij}\) is the GPU allocation for step \(s_{ij}\).

\paragraph{Offline Profiling for Cost Model.}  
To make the optimization tractable, \name profiles execution times offline. 
For every step type $s_{ij}$ and GPU count $k \in \{1,2,4,\dots,N\}$, we measure the actual execution time $T_{ij}(k)$. 
From this, we derive the GPU hour $k \times T_{ij}(k)$ and store it in a lookup table. At runtime, \name simply enumerates candidate GPU assignments using these pre-profiled values. 

\begin{figure}[t]
    \centering
    \includegraphics[width=\linewidth]{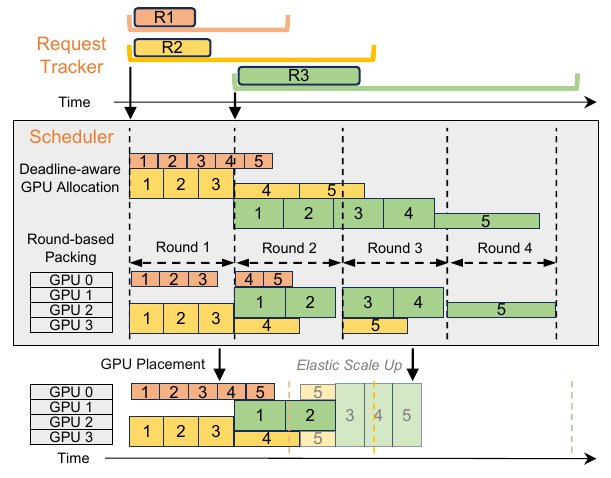}
    \caption{Illustration of \name's scheduling process. The progression is shown from top to bottom: each row represents an intermediate scheduling step, while the final row shows the actual GPU allocation decision. Time is fixed across rows.}
    \label{fig:scheduling}
\end{figure}

The above process aims to assign each request the minimum number of GPUs required to meet its deadline while minimizing the total GPU hours. 
Figure~\ref{fig:scheduling} illustrates this process with a concrete example: three requests (R1--R3), each with five steps, arrive over time. R1 has a small resolution (e.g., 256) and is fixed at SP=1 since higher parallelism would reduce efficiency (see Figure~\ref{fig:e2e_scaling_efficiency}). For R2 and R3, \name identifies GPU allocations with two parallelism degrees that just meet their deadlines while minimizing overall GPU usage.
The GPU allocations produced by this selection serve as the input to the subsequent request packing stage, where \name schedules requests across GPUs to maximize goodput.


\subsubsection{Request Packing}

The objective of scheduling is to maximize the number of requests that complete before their deadlines. 
To make the problem tractable, we approximate it by minimizing the number of requests that become \emph{definitely late}---those that cannot meet their deadlines even under maximal parallelism if not advanced in the current round. 
Deadline-aware GPU allocation determines the minimal GPU allocations needed for each request to meet its deadline while minimizing GPU hours. 
This makes it possible to pack more requests into each round and thereby reduce the number that would otherwise be definitely late.

After deadline-aware GPU allocation, each request $req_i$ is described by a set of allocations $(s_i^m, A_i^m)$, where $s_i^m$ is the number of steps executed with allocation $A_i^m$, and per-step times $T_i(A_i^m)$ are obtained from the cost model. 
To schedule requests across $N$ GPUs, \name divides time into \emph{rounds} of fixed duration $\tau$, which serves as the scheduling granularity. 
The choice of $\tau$ balances overhead and responsiveness: shorter rounds allow finer-grained preemption and more adaptive scheduling, while longer rounds reduce overhead but make scheduling coarser. 

At the beginning of each round $r$ (time $t_r$), the scheduler considers all pending requests and their allocations, and decides which to place within the $N$ GPUs. Within a round of duration $\tau$, if GPU allocation $m$ of request $i$ is chosen, the number of steps that can complete is
\[
q_i^m = \min\!\Big\{ s_i^m,\ \big\lfloor \tfrac{\tau}{T_i(A_i^m)} \big\rfloor \Big\}.
\]
Options with $q_i^m=0$ are discarded to avoid wasting resources. Choosing option $o \in \{\textsf{none},1,2,\dots\}$ updates the remaining steps as
\[
\tilde{s}_i^m(o) = s_i^m - \mathbb{I}[o=m] \, q_i^m,
\]
clipped at zero, where $\mathbb{I}[o=m]$ equals $1$ if $o=m$ and $0$ otherwise. The next round begins at $t_{r+1}=t_r+\tau$.

To decide which requests must be scheduled \emph{now}, we identify those that would become \emph{definitely late} at $t_{r+1}$ if not advanced in this round. Using the fastest possible step time $T_i^{\min}=\min_{k\in {\{1, 2, 4, \ldots, N\}}} T_i(k)$, we define the \emph{residual completion time lower bound} under option $o$ as
\[
\mathrm{LB}_i(o) = \Big(\sum_m \tilde{s}_i^m(o)\Big)\,T_i^{\min},
\]
where $\tilde{s}_i^m(o)$ is the updated step count. A request survives only if
\[
t_{r+1}+\mathrm{LB}_i(o)\le D_i.
\]

Each option $o$ consumes $w_i(o)$ GPUs: $w_i(\textsf{none})=0$, $w_i(m)=A_i^m$. The per-round scheduling problem is therefore to select at most one option per request, with total GPU $\le N$, maximizing the number of requests that survive to the next round. For requests that have already missed their deadlines, we assign at most one GPU in a best-effort manner without impacting other requests, and scale them up later if idle GPUs become available.
By anchoring scheduling decisions on the round duration $\tau$, \name balances preemption overhead and responsiveness, while ensuring urgent requests receive priority.



\begin{algorithm}[t]
\caption{DP Round Scheduler}
\label{alg:dp_round}
\SetKwInOut{Input}{Input}
\SetKwInOut{Output}{Output}
\Input{Pending requests $R$ with $\{(s_i^m, A_i^m)\}_{m\in\mathcal{M}_i}$ and $T_i(\cdot)$; capacity $N$; round length $\tau$; current time $t_r$}
\Output{Selected plan}
$t_{r+1} \leftarrow t_r + \tau$\;
\ForEach{$i \in R$}{
    \ForEach{$m \in \mathcal{M}_i$}{
        $q_i^m \leftarrow \min\{s_i^m, \lfloor \tau / T_i(A_i^m) \rfloor\}$\;
    }
    $T_i^{\min} \leftarrow \min_{k\in K} T_i(k)$\;
    $\mathcal{O}_i \leftarrow \{\textsf{none}\} \cup \{\,m \in \mathcal{M}_i \mid q_i^m>0 \wedge A_i^m \le N\,\}$\;
    \ForEach{$o \in \mathcal{O}_i$}{
        \ForEach{$m \in \mathcal{M}_i$}{
            $\tilde{s}_i^m(o) \leftarrow s_i^m - \mathbb{I}[o=m]\cdot q_i^m$ \label{alg:line:steps}\;
        }
        $\mathrm{LB}_i(o) \leftarrow \big(\sum_{m \in \mathcal{M}_i} \tilde{s}_i^m(o)\big)\,T_i^{\min}$ \label{alg:line:lb}\;
        $\mathrm{sv}_i(o) \leftarrow \mathbb{I}[\,t_{r+1}+\mathrm{LB}_i(o) \le D_i\,]$\;
        $w_i(o) \leftarrow 0$ if $o=\textsf{none}$ else $A_i^{o}$ \label{alg:line:width}\;
    }
}
\BlankLine
Initialize $\text{dp}[0..N] \leftarrow -\infty$, $\text{dp}[0]\leftarrow 0$\;
\ForEach{$i \in R$}{
    $\text{next}[0..N] \leftarrow \text{dp}$\;
    \For{$c=0$ \KwTo $N$}{
        \ForEach{$o \in \mathcal{O}_i$}{
            \If{$w_i(o) \le c$}{
                $\text{next}[c] \leftarrow \max\{\text{next}[c],\ \text{dp}[c-w_i(o)] + \mathrm{sv}_i(o)\}$\;
            }
        }
    }
    $\text{dp} \leftarrow \text{next}$\;
}
$c^\star \leftarrow \arg\max_{c} \text{dp}[c]$\;
\Return plan reconstructed from back-pointers at $c^\star$\;
\end{algorithm}

\paragraph{Dynamic Programming.} Naively enumerating all per-request options \(\mathcal{O}_i\) for feasible packings within a round is exponential in the number of requests and quickly becomes intractable. 
We observe that the per-round decision has the \emph{group-knapsack} structure: for each request \(i\) (a group), we must choose at most one option \(o\) (run one of its GPU allocation this round or \textsf{none}), each option consumes width (GPUs) and yields a binary “survival” value indicating whether the request is \emph{not definitely late} at the next round start. 
This lets us replace exhaustive search with a dynamic program (DP) that maximizes the number of surviving requests under the round capacity \(N\).

Concretely, the DP state \(\mathrm{dp}[c]\) stores, after processing the first \(i\) requests, the maximum number of surviving requests achievable with exactly capacity \(c\in\{0,\dots,N\}\) consumed in the current round. 
For request \(i\), we build its option set \(\mathcal{O}_i\) once (group constraint): \textsf{none} (consume zero GPUs, no progress) and one option per allocation \(m\) that can make progress in this round, i.e., \(q_i^m=\big\lfloor \tau/T_i(A_i^m)\big\rfloor>0\) and \(A_i^m\le N\). 
For each option \(o\in\mathcal{O}_i\), we compute:
\begin{enumerate}
    \item \textbf{Line~\ref{alg:line:steps}:} the updated remaining steps \(\tilde{s}_i^m(o)\).
    \item \textbf{Line~\ref{alg:line:lb}:} a conservative lower bound \(\mathrm{LB}_i(o)\) on the residual processing time from \(t_{r+1}=t_r+\tau\).
    \item \textbf{Line~\ref{alg:line:width}:} its width \(w_i(o)\) (\(0\) for \textsf{none}, \(A_i^m\) for allocation \(m\)). 
\end{enumerate}

We then set the survival indicator \(\mathrm{sv}_i(o)=\mathbb{I}[\,t_{r+1}+\mathrm{LB}_i(o)\le D_i\,]\).
The DP transition iterates options once per request (respecting the group constraint) and, for each capacity \(c\), admits only options with \(w_i(o)\le c\) (respecting the capacity constraint):
\[
\mathrm{next}[c]\leftarrow\max\big\{\mathrm{next}[c],\ \mathrm{dp}[c-w_i(o)]+\mathrm{sv}_i(o)\big\}.
\]
Using a rolling array yields \(O(N)\) space. Since each request contributes at most \(|\mathcal{O}_i|\) options, DP runs in \(O(RN)\) time and \(O(N)\) space per round (rolling array), which is tractable even at millisecond-scale rounds for moderate \(N\). This is orders of magnitude cheaper than enumerating all feasible packing combinations.


\paragraph{Round Duration.} 
Algorithm~\ref{alg:dp_round} schedules in fixed-length rounds of duration \(\tau\). 
The choice of \(\tau\) balances two factors: short rounds reduce admission delay for new requests but increase scheduling frequency, while long rounds amortize scheduling cost but risk larger queueing delay and deadline misses. 
For a given GPU configuration (e.g., NVIDIA H100), \name adapts \(\tau\) to the step execution times of requests across different resolutions, so that requests with heterogeneous step lengths can finish around the same round boundary. 
This minimizes idle bubbles while keeping \(\tau\) short enough to avoid excessive queueing delay. 
In practice, we determine \(\tau\) by the \emph{step granularity}, which means each round executes multiple diffusion steps.
We will further discuss the impact of round duration in the evaluation section (\S\ref{sec:eval:sensitivity}).

\subsubsection{Efficient GPU Placement and Allocation}
In the round-based framework (Algorithm~\ref{alg:dp_round}), \name improves efficiency via two complementary steps: \emph{placement preservation} and \emph{work-conserving elastic scale-up}, illustrated in Figure~\ref{fig:scheduling}. 
First, to avoid idle bubbles between rounds, \name adopts a placement-aware policy: requests continue on the same GPUs across consecutive rounds whenever possible. This eliminates state-transfer delays and ensures immediate progress at round boundaries. 

Second, any GPUs left idle after placement are reclaimed through a work-conserving elastic scale-up policy. Requests with sufficient remaining steps are granted additional GPUs if $T_i(k_i') < T_i(k_i)$, prioritizing those that benefit most from parallelism. This ensures no GPU remains unused within a round, reducing future load and improving deadline satisfaction. Together, placement preservation minimizes inter-round stalls, while elastic scale-up guarantees work-conserving allocation within each round.

\section{Implementation}\label{sec:implementation}

\name is implemented in 5,033 lines of Python and C++ code. 
We reuse components from existing solutions, including the sequence parallelism engine from xDiT \cite{xDiT-arxiv24}, async logic from vLLM \cite{PagedAttention-sosp23}, and process launcher from MuxServe \cite{MuxServe-icml24} and SGLang \cite{SGLang-neurips24}. 

\paragraph{Scheduler.} 
The scheduler's core decision loop is implemented in C++ and exposed via lightweight bindings, achieving millisecond-level control-plane latency.

\paragraph{VAE Decoder Sequential Execution.}
The VAE decoder imposes a large activation-memory footprint at high resolutions and batch sizes, whereas its wall-clock cost is very small relative to diffusion steps.
Accordingly, we adopt sequential per-request decoding to bound peak memory by avoiding concurrent decoder activations across a batch. 
Because the decoder is largely off the critical path, this design does not increase end-to-end latency. 
The reduced peak usage also increases headroom for model state and communication buffers, lowering the risk of out-of-memory failures under mixed workloads.

\paragraph{Communication Process Groups Warmup.}
We pre-create process groups for all relevant combinations of devices (e.g., \(\binom{8}{k}\) groups for degrees \(k \in \{1,\dots,8\}\)). 
Creating the group itself is lightweight and does not materially consume GPU memory. 
However, the \emph{first} invocation on a group initializes NCCL \cite{nccl2022} channels and allocates persistent device buffers for subsequent collectives. 
Proactively warming \emph{every} group therefore inflates memory usage and can exceed available HBM. 
To balance startup latency and memory footprint, we warm only a compact set of commonly used, overlapping groups (e.g., [0,1,2,3], [0,2,3,4]) and defer others to on-demand warmup. 
Empirically, this strategy preserves performance while maintaining low peak memory.

\paragraph{Latent Transfer.}
Because \name executes at step granularity, intermediate latents and lightweight metadata must be handed off across GPU groups. 
We provide a Future-like abstraction for latents that enables asynchronous, non-blocking transfer between steps. 
Latent tensors are compact (in the compressed latent space), so transfer overhead is negligible; consequently, the scheduler excludes latent-transfer time from deadline accounting.
We quantify this overhead in Section~\ref{sec:eval:sensitivity} and show it remains below 0.05\% of per-step latency across all configurations.

\paragraph{Selective Continuous Batching.}
Batching in diffusion inference is only effective for identical, small-resolution requests that would otherwise underutilize GPUs. This creates a throughput-latency trade-off. Our scheduler employs a selective, step-level batching strategy that only groups requests if their SLOs are not compromised, thus improving resource utilization without harming latency.


\section{Evaluation}
\label{sec:eval}

We evaluate \name against state-of-the-art baselines across diverse workloads. Key findings:

\begin{denseitemize}
    \item \name outperforms baselines by up to 32\% across all resolutions (\S\ref{sec:eval:e2e}).
    
    \item \name is robust to bursty arrivals and adapts to changing resolution mixes (\S\ref{sec:eval:deepdive}).
    
    \item Sensitivity analysis confirms \name's advantage holds across varying arrival rates, step granularities, and homogeneous workloads (\S\ref{sec:eval:sensitivity}).

    \item Ablation studies show that GPU placement preservation and elastic scale-up are crucial to \name's performance (\S\ref{sec:eval:ablation_study}).
\end{denseitemize}

\subsection{Methodology}
\label{sec:eval:methodology}

\paragraph{Testbed.} 
We conduct experiments on two GPU clusters. 
The first comprises nodes with 8 NVIDIA H100-80GB HBM3 GPUs interconnected via NVLink 4.0 (900 GB/s inter-GPU bandwidth). 
The second features nodes with 4 NVIDIA A40-48GB GPUs connected in pairs via NVLink and interfaced to the host via PCIe 4.0. 
Our software environment is based on NVIDIA's NGC container with CUDA 12.5, NCCL 2.22.3 \cite{nccl2022}, PyTorch 2.4.0 \cite{zhao2023pytorch}, and xDiT \cite{xDiT-arxiv24} (git-hash 8f4b9d30).

\paragraph{Models and Metrics.} 
We select \emph{FLUX.1-dev} \cite{flux1dev2024} and \emph{Stable Diffusion 3 Medium} (SD3) \cite{sd3} as representative models, evaluating them on H100 and A40 clusters, respectively.
We report SLO Attainment Ratio (SAR; fraction of requests finishing within SLO) as our primary metric and plot end-to-end latency CDFs to show the latency distribution.



\paragraph{Baselines.}
We compare \name against:

\begin{itemize}[leftmargin=*,itemsep=2pt]
    \item \textbf{xDiT (SP=1/2/4/8).}
    Fixed sequence parallelism degree; each request uses a constant number of GPUs. 

    \item \textbf{Resolution-Specific SP (RSSP).}
    Selects the best SP degree per resolution via offline profiling: SP=1 for $256\times256$ and $512\times512$, SP=2 for $1024\times1024$, and SP=8 for $2048\times2048$.
    Represents an oracle static configuration.
\end{itemize}

\paragraph{SLO Settings.} 
We adopt resolution-specific latency targets grounded in user-perceived responsiveness. 
Prior research \cite{abbas2022understanding} reports that 63\% of users prefer a maximum response delay of 5 seconds in interactive settings. Accordingly, we cap the target at 1.5 seconds for small images and set an upper bound of 5.0 seconds for the largest resolution: $(256,256)$ = 1.5 s, $(512,512)$ = 2.0 s, $(1024,1024)$ = 3.0 s, and $(2048,2048)$ = 5.0 s.
We sweep SLO Scale from $1.0\times$ to $1.5\times$ relative to each resolution's baseline.

\paragraph{Workload and Dataset.} 
We sample 300 prompts from DiffusionDB \cite{diffusionDB} to generate requests.
By default, requests arrive as a Poisson process at 12 requests/minute.

We consider two resolution mixes: 
\begin{itemize}[leftmargin=*,itemsep=2pt]
    \item \emph{Uniform}: equal number of requests across resolutions \{256, 512, 1024, 2048\}.

    \item \emph{Skewed}: resolutions sampled with exponential weight over latent length, \(p_i \propto \exp(\alpha \cdot L_i/L_{\max})\), with \(\alpha=1.0\) and \(L_i=(H_i\cdot W_i)/16^2\), biasing toward larger resolutions.
\end{itemize}


\subsection{End-to-End Performance}
\label{sec:eval:e2e}

\begin{figure}[t]
    \centering
    \begin{subfigure}[b]{\columnwidth}
        \centering
        \includegraphics[width=\linewidth]{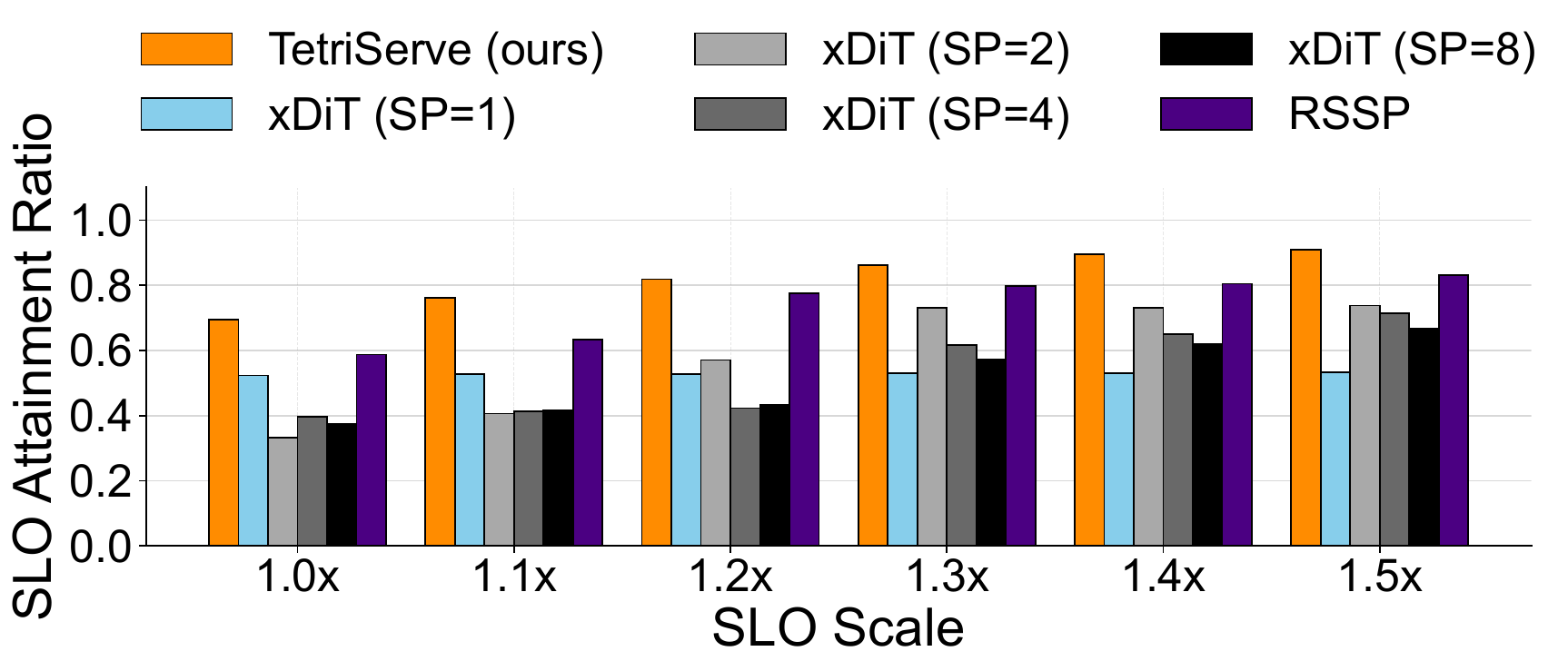}
        \caption{SLO Attainment Ratio (SAR) of Uniform Workload}
        \label{fig:sar_uniform:a}
    \end{subfigure}
    
    \vspace{0.5em}

    \begin{subfigure}[b]{0.48\columnwidth}
        \centering
        \includegraphics[width=\linewidth]{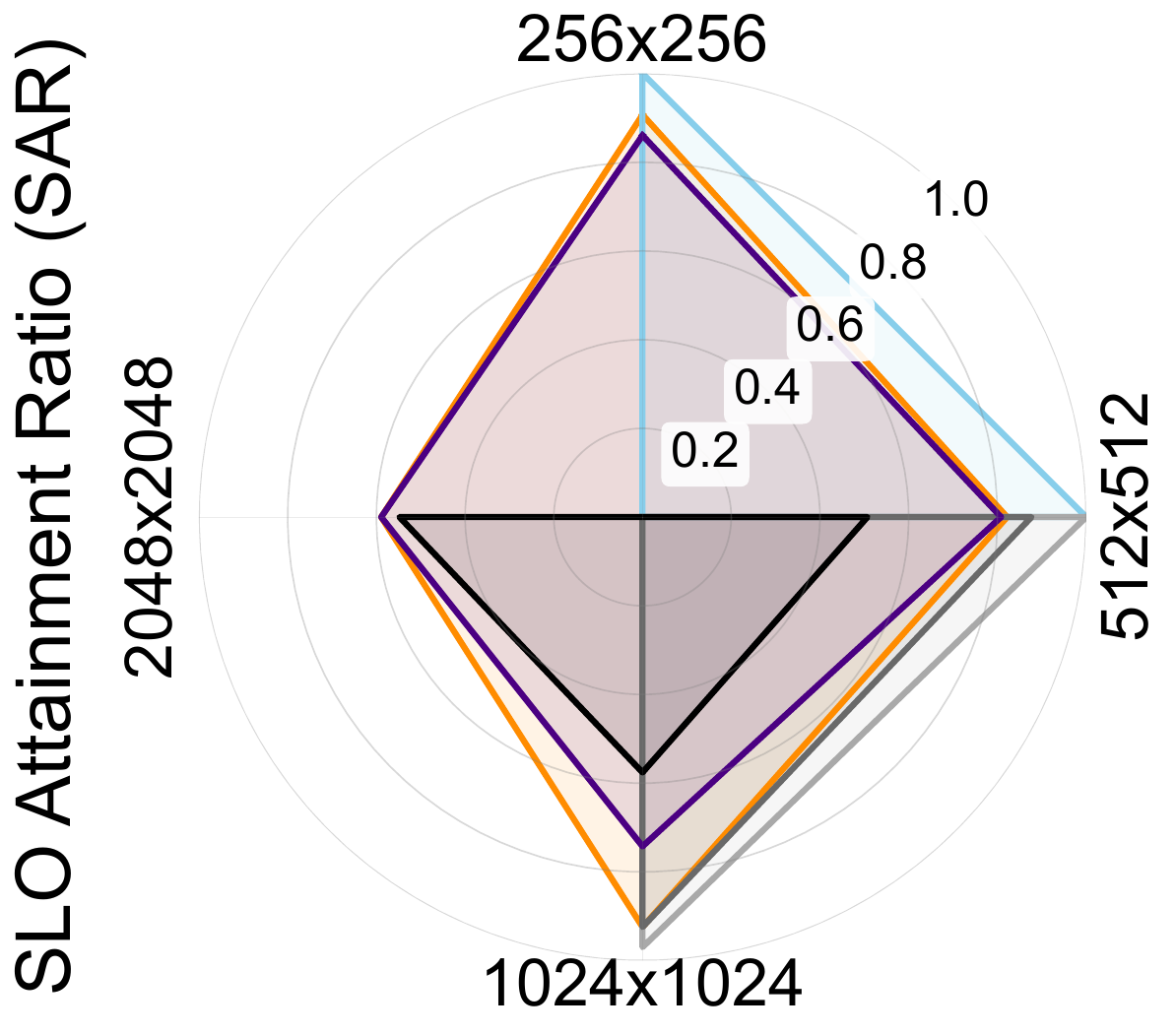}
        \caption{Uniform, SLO Scale=1.0$\times$}
        \label{fig:sar_uniform:b}
    \end{subfigure}
    \hfill
    \begin{subfigure}[b]{0.48\columnwidth}
        \centering
        \includegraphics[width=\linewidth]{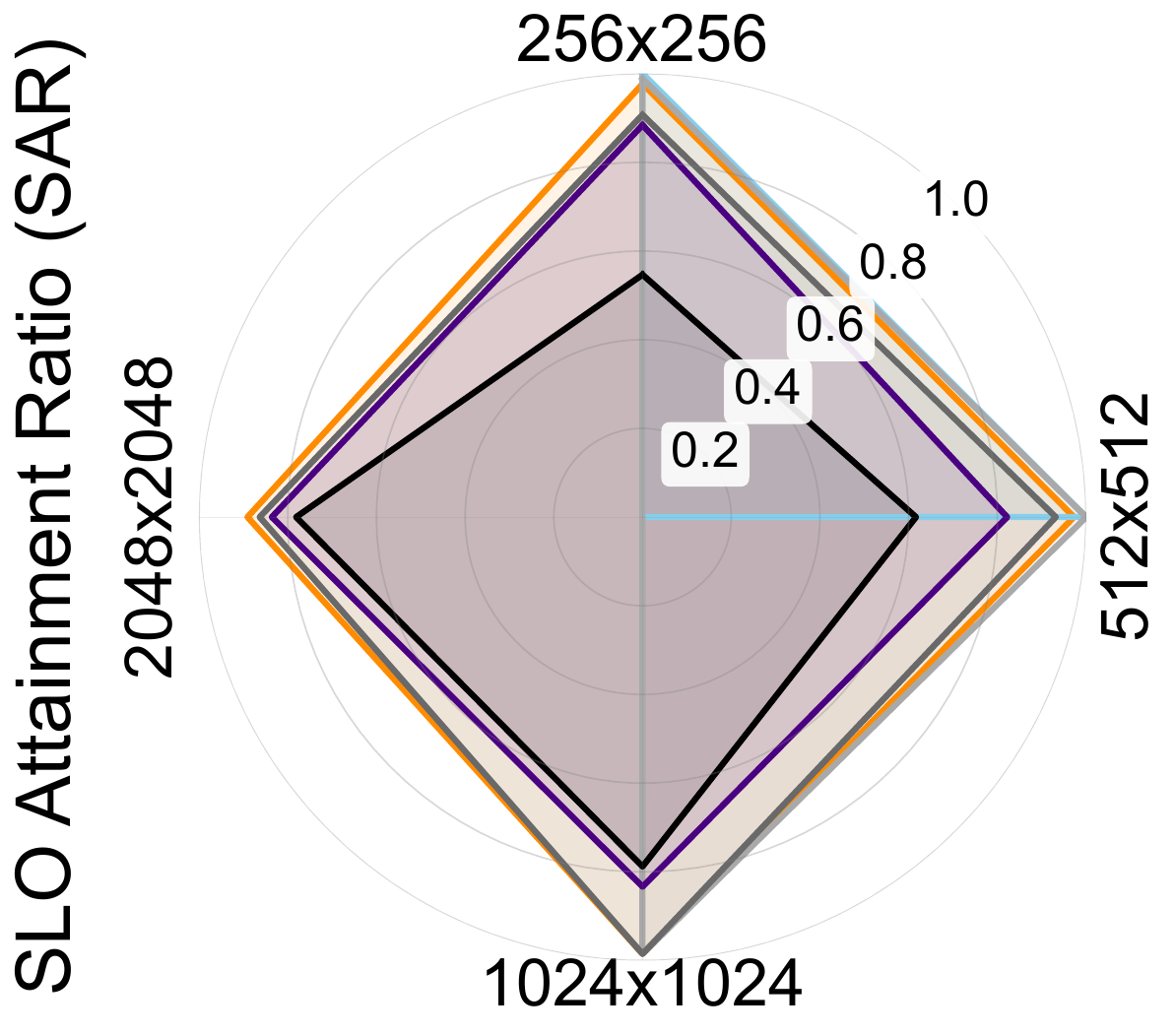}
        \caption{Uniform, SLO Scale=1.5$\times$}
        \label{fig:sar_uniform:c}
    \end{subfigure}
    \caption{End-to-end performance on the Uniform workload at 12 req/min. \textbf{(Top)} \name achieves the highest SLO Attainment Ratio (SAR) across all SLO scales. \textbf{(Bottom)} The spider plots show that xDiT variants only perform well for specific resolutions, \name delivers high SAR across all resolutions no matter tight or loose SLO Setting.}
    \label{fig:sar_uniform}
\end{figure}

\begin{figure}[t]
    \centering
    \begin{subfigure}[b]{\columnwidth}
        \centering
        \includegraphics[width=\linewidth]{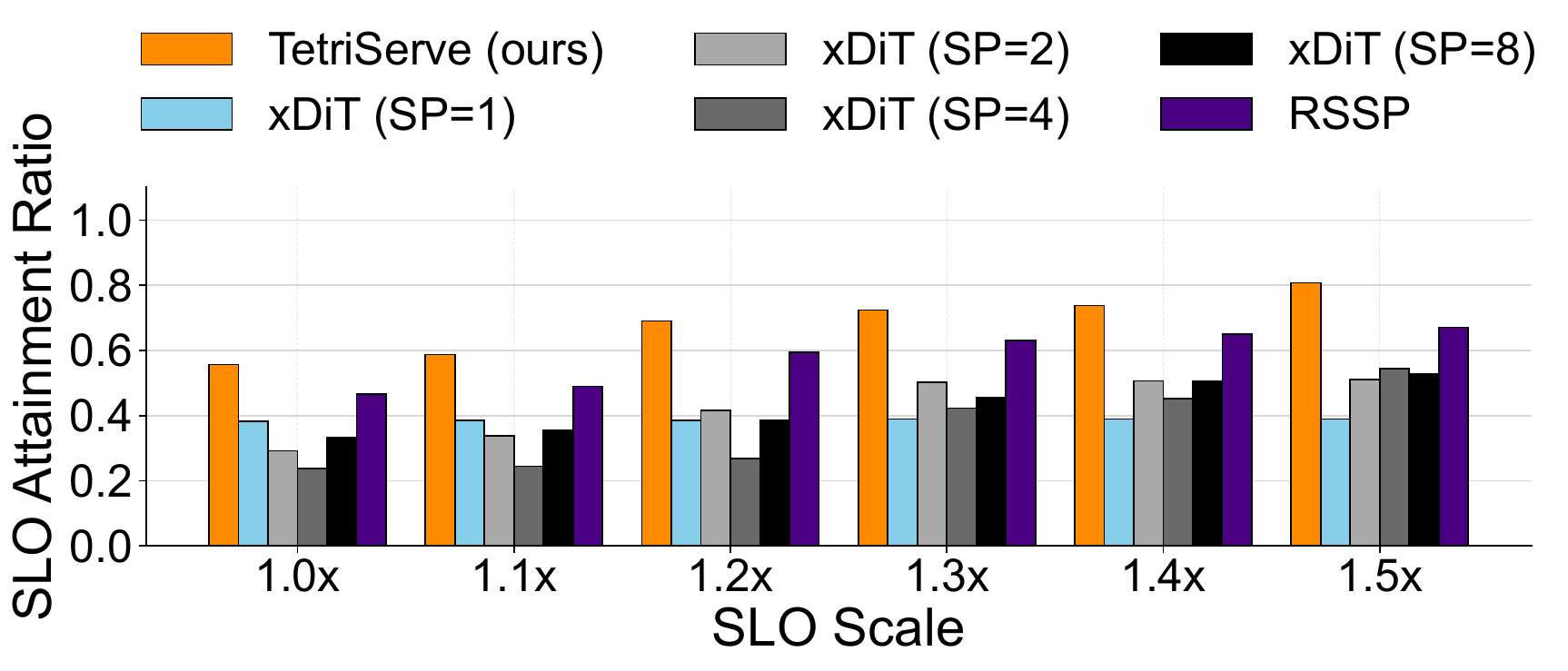}
        \caption{SAR of Skewed Workload}
        \label{fig:sar_skewed:a}
    \end{subfigure}
    
    \vspace{0.5em}

    \begin{subfigure}[b]{0.48\columnwidth}
        \centering
        \includegraphics[width=\linewidth]{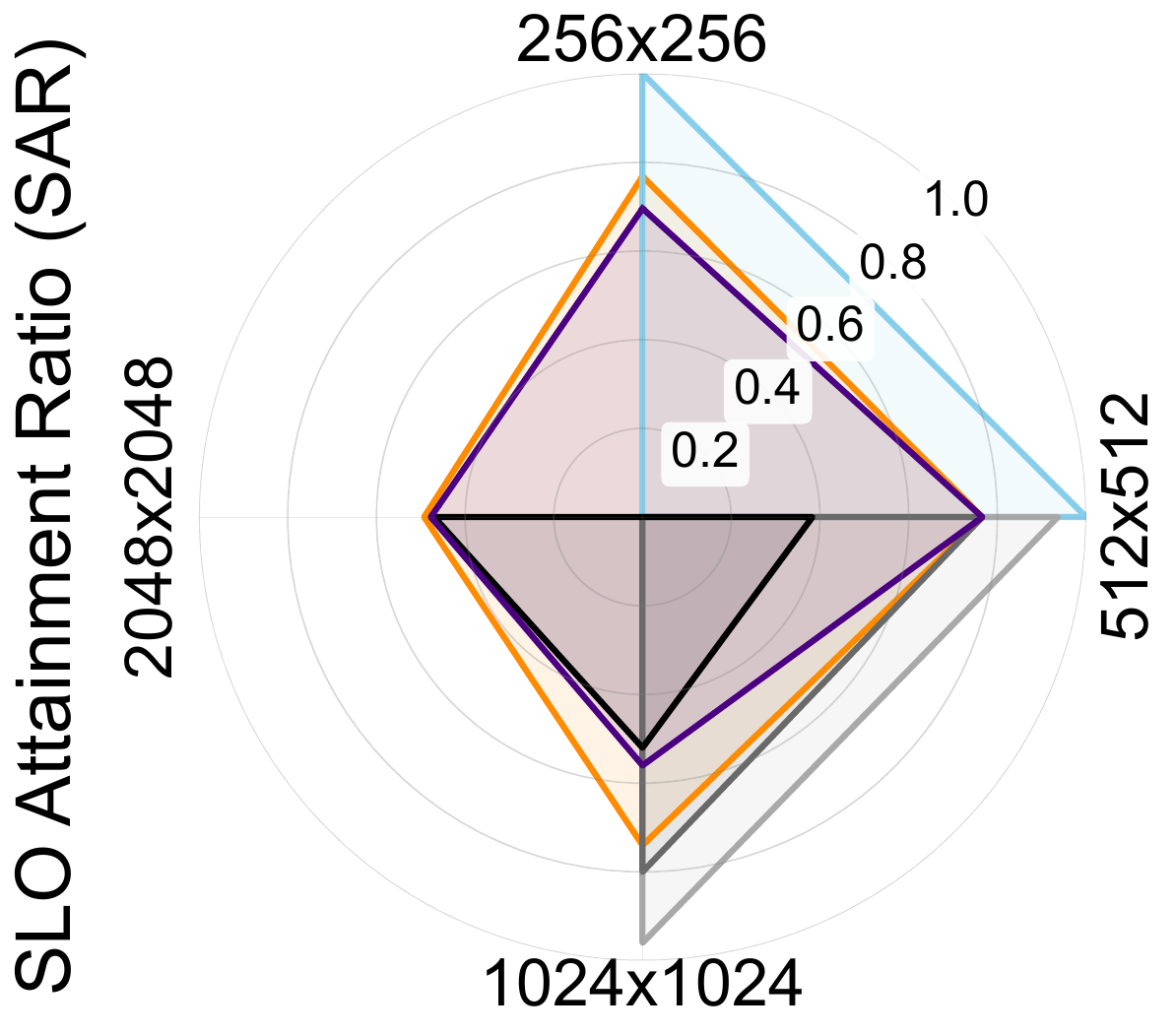}
        \caption{Skewed, SLO Scale=1.0$\times$}
        \label{fig:sar_skewed:b}
    \end{subfigure}
    \hfill
    \begin{subfigure}[b]{0.48\columnwidth}
        \centering
        \includegraphics[width=\linewidth]{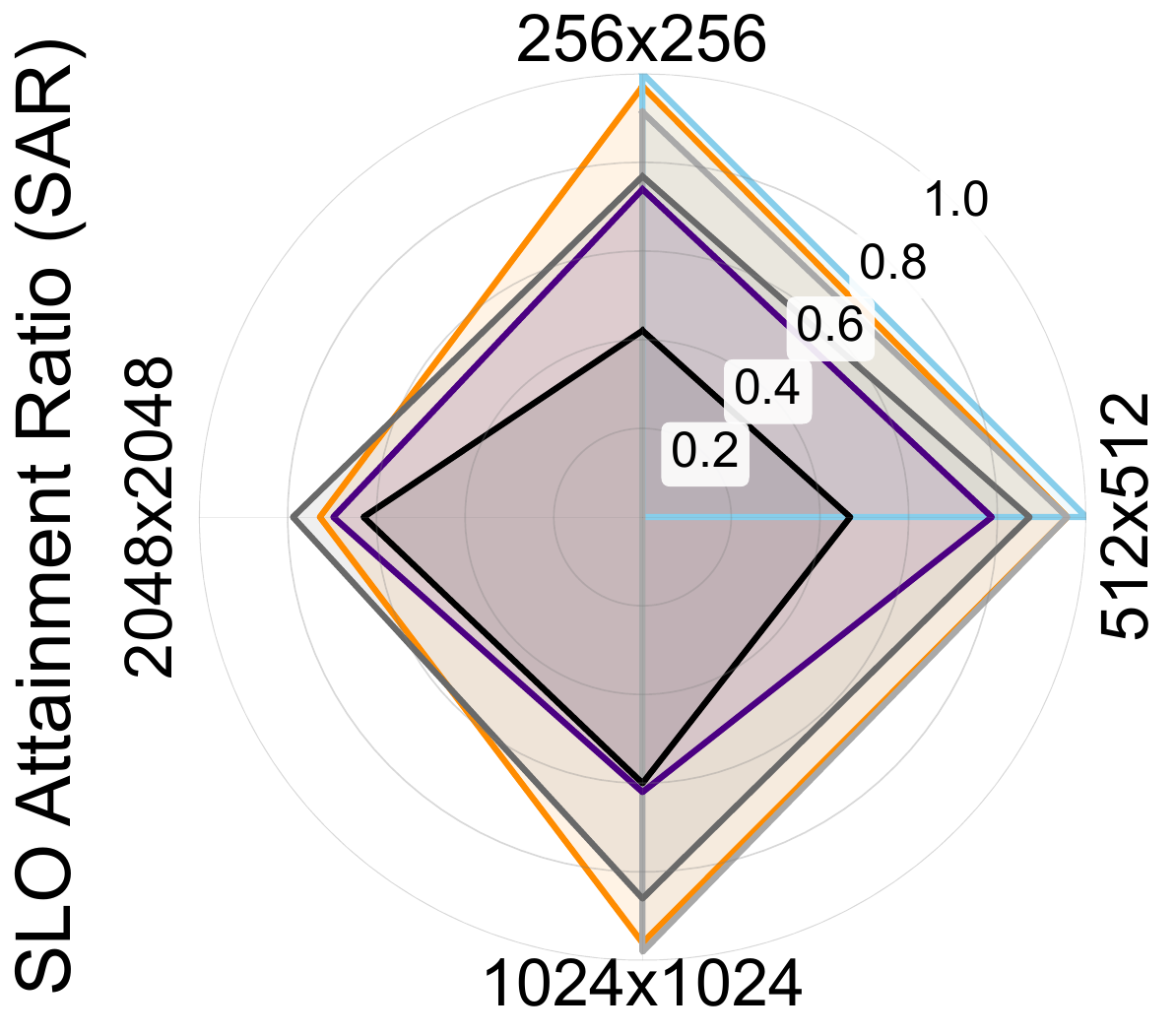}
        \caption{Skewed, SLO Scale=1.5$\times$}
        \label{fig:sar_skewed:c}
    \end{subfigure}
    \caption{End-to-end performance on the Skewed workload at 12 req/min. \textbf{(Top)} \name again achieves the highest SLO Attainment Ratio (SAR) across all SLO scales. \textbf{(Bottom)} The spider plots confirm that \name's adaptive parallelism provides robust performance across all resolutions, even in a workload dominated by large images}
    \label{fig:sar_skewed}
\end{figure}

\paragraph{\name Improves SAR.}
Figures~\ref{fig:sar_uniform} and~\ref{fig:sar_skewed} show the end-to-end SLO Attainment Ratio (SAR) of \name compared to fixed-parallelism baselines for FLUX on H100s for both the Uniform and Skewed workload mixes at an arrival rate of 12 requests per minute. 
As shown in Figures~\ref{fig:sar_uniform:a} and~\ref{fig:sar_skewed:a}, \name consistently achieves the highest SAR across all SLO scales and both workload distributions. This demonstrates the effectiveness of its step-level parallelism control and request packing, which allow it to dynamically adapt to the workload and outperform the rigid strategies of the baselines. 

On average, \name outperforms the best fixed parallelism strategy by 10\% for the Uniform mix and 15\% for the Skewed mix.
The performance gap is particularly pronounced at tighter SLOs. For instance, with an SLO scale of 1.1$\times$ in the Uniform mix, \name outperforms the best baseline by 28\%. Similarly, in the Skewed mix with a 1.2$\times$ SLO scale, \name's SAR is 32\% higher than the best-performing fixed strategy. 

Notably, this advantage holds even when compared against RSSP, a strong per-resolution baseline that selects the best fixed parallelism degree for each input resolution.
Despite this, RSSP remains fundamentally limited by its lack of deadline awareness and runtime adaptation, whereas \name dynamically adjusts parallelism at the step level to meet per-request SLOs.
This highlights \name's superior performance under challenging, tightly constrained Workloads.

\paragraph{\name Benefits All Resolutions.}
\name's strength lies in its ability to deliver high SAR across all request resolutions, unlike fixed strategies that only excel at specific ones. 
The spider plots in the bottom row of Figures~\ref{fig:sar_uniform} and~\ref{fig:sar_skewed} break down SAR by resolution. With a relaxed SLO of 1.5$\times$ (Figures~\ref{fig:sar_uniform:c} and~\ref{fig:sar_skewed:c}), \name achieves near-perfect SAR across all resolutions for both workload mixes, consistently outperforming all xDiT baselines. 
Under the tightest SLO of 1.0$\times$ (Figures~\ref{fig:sar_uniform:b} and~\ref{fig:sar_skewed:b}), \name provides the best overall performance. 
While some fixed-parallelism strategies may marginally outperform \name on a single resolution (e.g., xDiT SP=1 on 256px), they perform poorly on others. In contrast, \name dynamically adapts its parallelism, providing high SAR across the entire spectrum of resolutions. 

Conceptually, RSSP is a restricted variant of \name in which the scheduler cannot adjust parallelism beyond a fixed configuration. Since RSSP explores only a subset of \name's decision space, it cannot exploit additional parallelism for deadline-critical requests, resulting in uniformly lower SAR across resolutions.
In contrast, \name avoids over parallelization for less urgent requests and prioritizes more GPU resources for more urgent requests, thus performing well on all resolutions.

\begin{figure}[t]
    \centering
    \includegraphics[width=0.92\columnwidth]{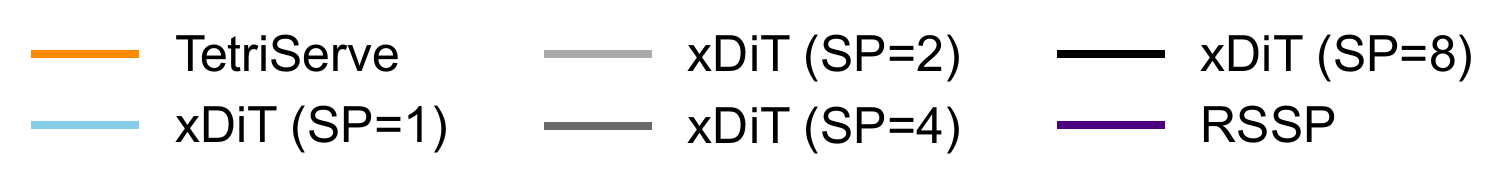}
    \vspace{-1em}
    \begin{subfigure}[b]{0.48\columnwidth}
        \centering
        \includegraphics[width=\linewidth]{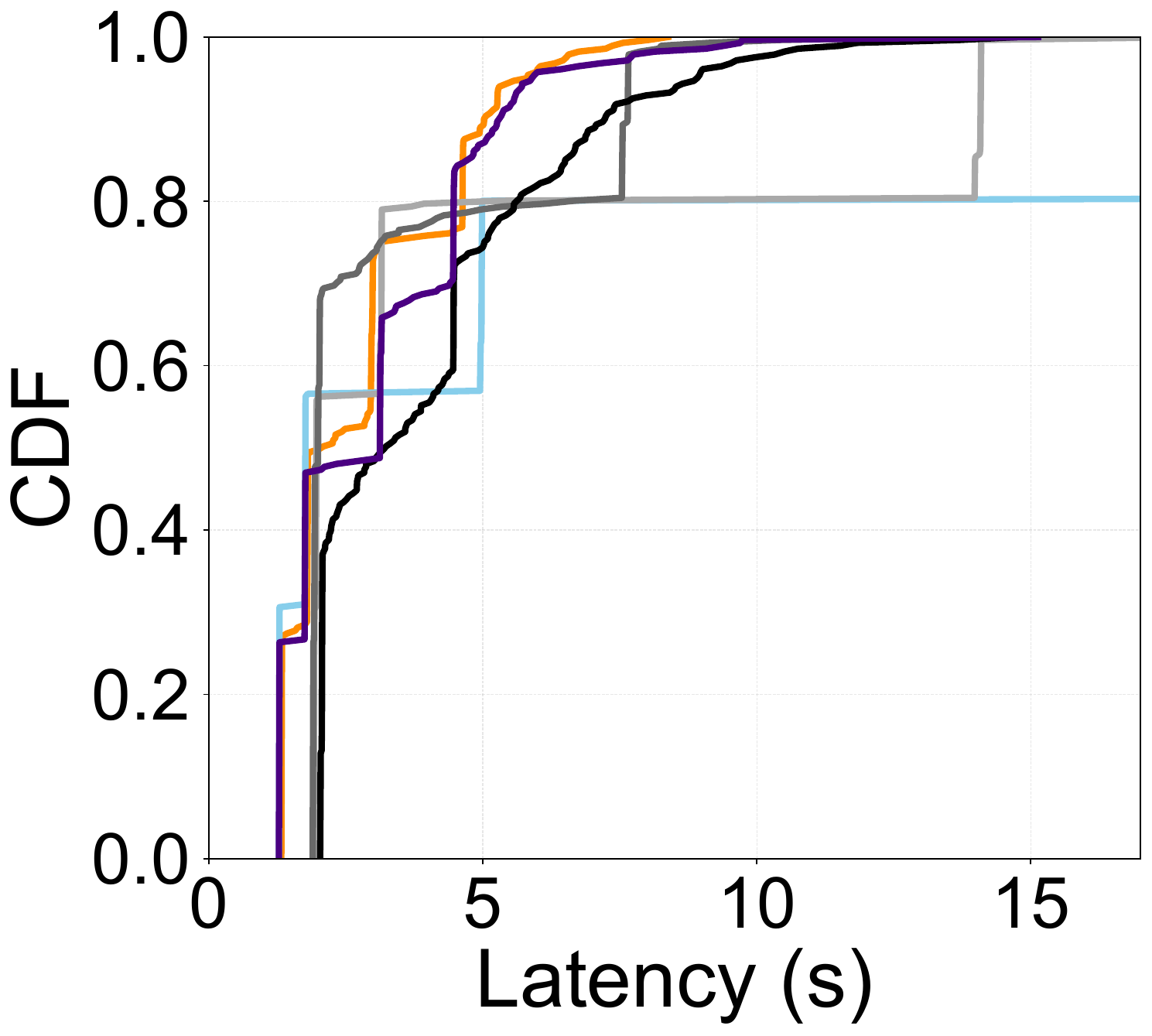}
        \caption{Uniform mix}
        \label{fig:cdf_latency:uniform}
    \end{subfigure}
    \hfill
    \begin{subfigure}[b]{0.48\columnwidth}
        \centering
        \includegraphics[width=\linewidth]{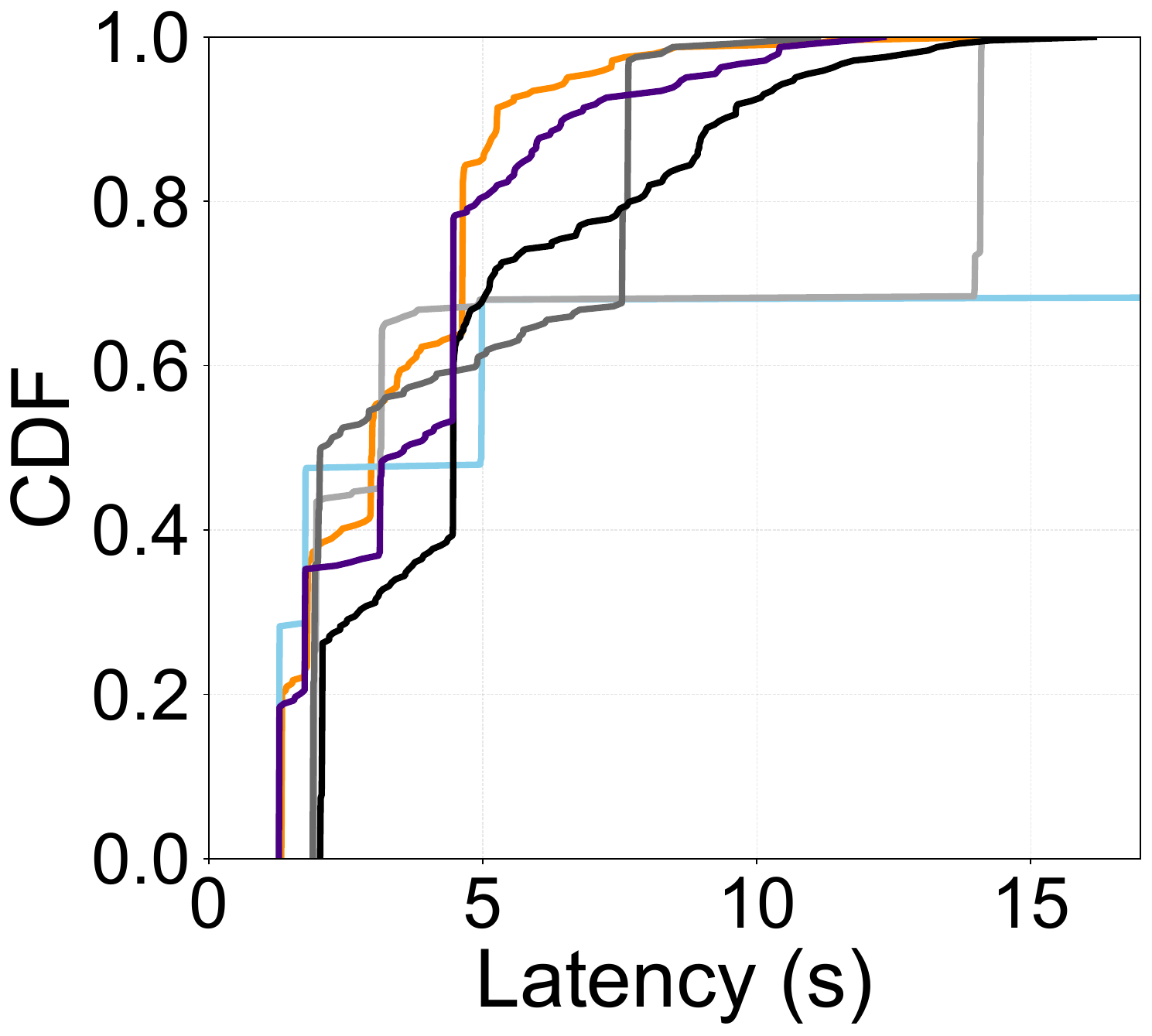}
        \caption{Skewed mix}
        \label{fig:cdf_latency:skewed}
    \end{subfigure}
    \caption{\textbf{End-to-end latency CDF under strict SLOs (FLUX on H100, SLO scale = 1.0$\times$).} \name shows more consistent and better tail latency distribution than other baselines under strict SLO settings. The x-axis is truncated at 17s for readability; the SP=1 baseline has a much heavier tail beyond this range.}
    \label{fig:cdf_latency}
\end{figure}

\paragraph{Tail Latency.} Figure~\ref{fig:cdf_latency} plots the CDF of end-to-end request latency under the tightest SLO setting (SLO scale = 1.0$\times$) for both the Uniform and Skewed mixes.
We compute the CDF over completed requests only, i.e., requests that finish execution at least once (those that miss the deadline and are dropped/timeout are excluded from the latency distribution). 
Across both workload mixes, \name produces a consistently more favorable tail distribution than fixed-parallelism baselines and RSSP. 
Compared to fixed SP baselines, \name shifts the latency distribution left and reaches high completion probability at lower latency, indicating that most served requests finish quickly even under strict deadlines.
Compared to RSSP, which restricts scheduling to a smaller decision space, \name further reduces tail latency by dynamically reallocating GPUs toward more urgent requests and avoiding over-parallelization on less critical ones.
Overall, these results show that \name improves not only SAR but also keep the steady long tail latency under tight SLO scale.

\paragraph{Compatibility with Cache-Based Diffusion Acceleration.} \name is orthogonal and compatible with cache-based diffusion acceleration techniques.
To demonstrate this, we integrate Nirvana~\cite{ApproximateCaching-nsdi24} into our system.
Nirvana accelerates diffusion inference by reusing intermediate denoising latents from prior requests. 
Each incoming prompt is embedded using CLIP~\cite{clip-cvpr21} and matched against a cache of previously served prompts. 
Based on prompt similarity, the system determines how many initial diffusion steps can be skipped, yielding an effective diffusion length of $N-k$ steps, where $k \in \{5,10,15,20,25\}$ and $N=50$ by default.
We warm up the cache using the first 10K requests and then maintain a fixed-size cache with LRU eviction for online requests.

\begin{table}[t]
    \centering
    \small
    \caption{\textbf{SAR with Nirvana Integration.} SLO Attainment Ratio (SAR) under uniform and skewed workload mixes (12~req/min, SLO Scale $=1.0\times$). \name combined with Nirvana~\cite{ApproximateCaching-nsdi24} achieves the highest SAR by jointly exploiting cache-based step reduction and adaptive GPU parallelism.}
    \label{tab:nirvana_sar}
    \begin{tabular}{lcccc}
    \toprule
    \textbf{Workload} & \textbf{RSSP} & \textbf{\name} & \textbf{RSSP} & \textbf{\name} \\
    & & & \textbf{+ Nirvana} & \textbf{+ Nirvana} \\
    \midrule
    Uniform & 0.32 & 0.42 & 0.77 & \textbf{0.88} \\
    Skewed  & 0.04 & 0.19 & 0.53 & \textbf{0.75} \\
    \bottomrule
    \end{tabular}
\end{table}

Table~\ref{tab:nirvana_sar} compares four configurations: RSSP, \name, RSSP combined with Nirvana, and \name combined with Nirvana, under both Uniform and Skewed mix workloads under the SLO Scale of 1.0$\times$.
While Nirvana alone substantially improves SLO attainment by reducing per-request computation, it does not address resource fragmentation caused by heterogeneous request resolutions.
By contrast, \name further improves SLO attainment by dynamically adjusting GPU parallelism to match the reduced and variable step counts introduced by caching.
As a result, the combined system achieves the highest SLO attainment across both mixes, confirming that cache-based step reduction and \name's scheduling operate on complementary and orthogonal dimensions.

\begin{figure}
    \centering
    \includegraphics[width=0.98\columnwidth]{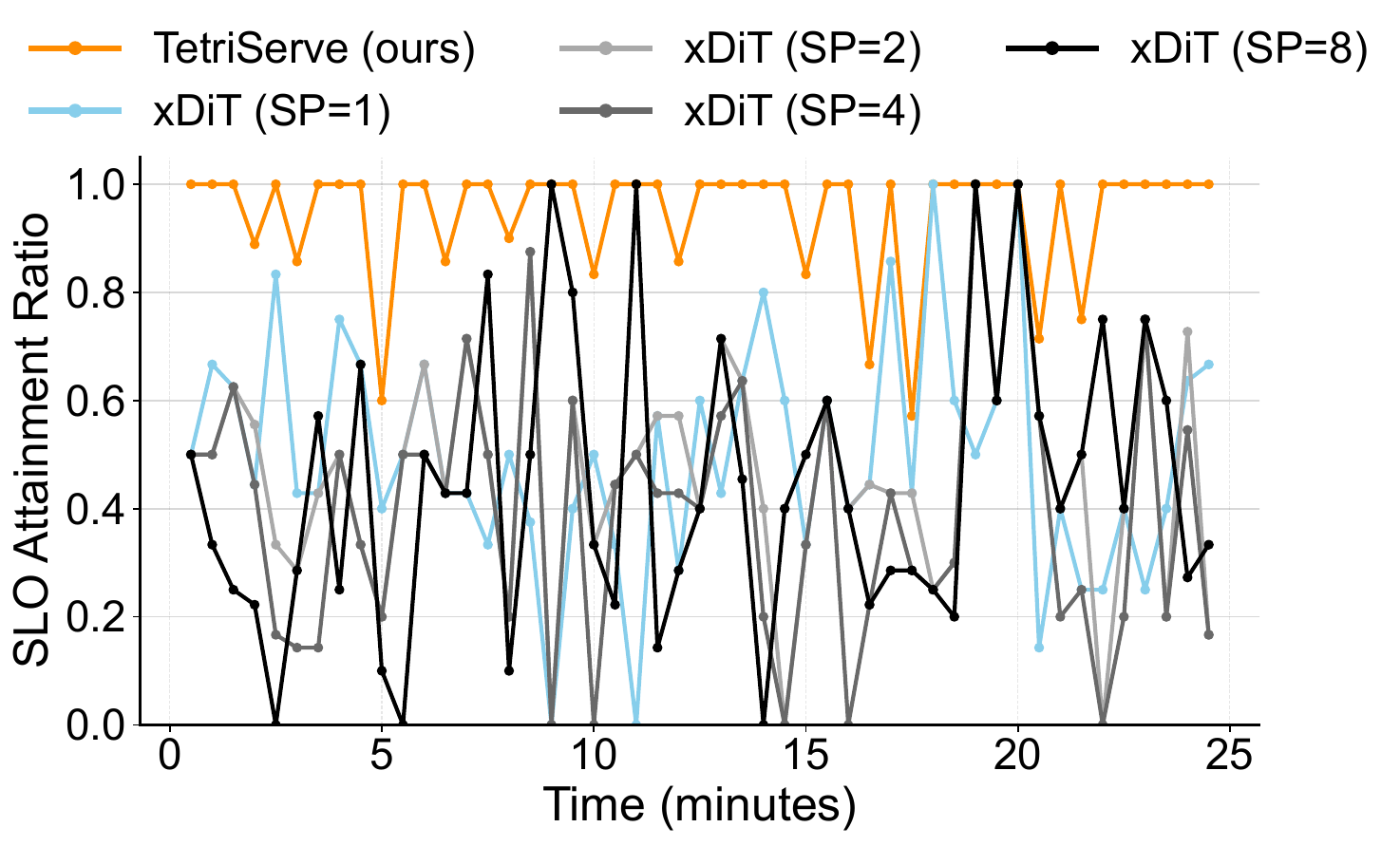}
    \caption{Performance stability under the Uniform workload at 12 req/min with a 1.5x SLO Scale. \name maintains a high and stable SLO Attainment Ratio (SAR) over time, which handles burstiness well.}
    \label{fig:sar_timeseries_even}
\end{figure}

\begin{figure}[t]
    \centering
    \includegraphics[width=0.95\columnwidth]{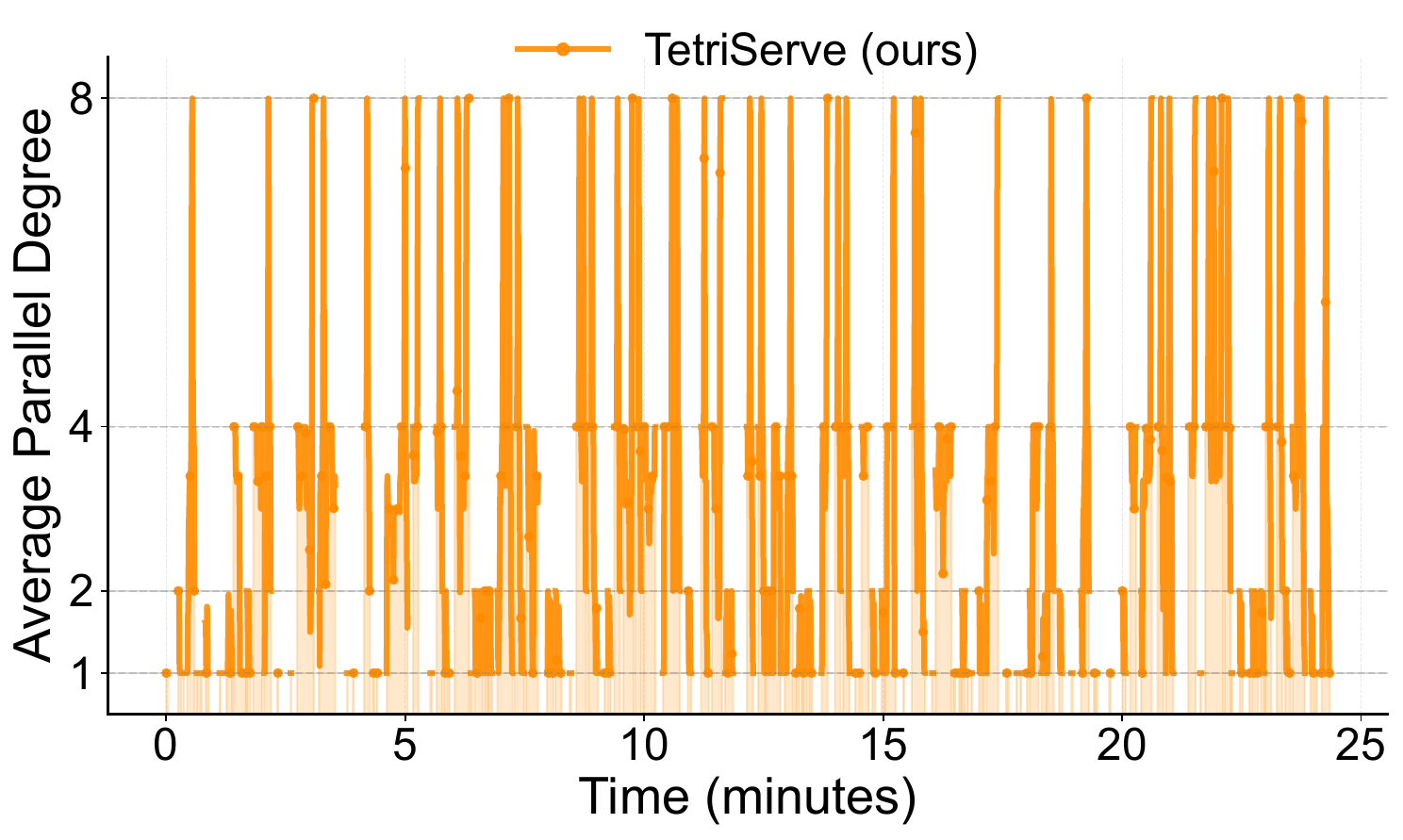}
    \caption{Average parallel degree of \name during serving under the Uniform workload (1.5$\times$ SLO Scale). \name dynamically adjusts sequence parallelism (SP) per request, assigning more GPUs to intensive requests (longer bars) to meet deadlines.}
    \label{fig:parallel_degree_timeseries_even}
\end{figure}

\subsection{Performance Stability under Bursty Traffic}
\label{sec:eval:deepdive}
\name maintains a high and stable SAR even under bursty arrival patterns, whereas fixed-parallelism approaches exhibit significant performance oscillations. 
For instance, Figure~\ref{fig:sar_timeseries_even} plots the SAR over time for the Uniform mix (12 req/min, SLO Scale=1.5$\times$). 
\name's SAR remains consistently high with low variance. 
In contrast, the fixed xDiT variants suffer from periodic drops in SAR, a result of utilization bubbles and subsequent queueing delays when bursty arrivals create contention. 

The key to \name's stability is its ability to adapt the degree of sequence parallelism (SP) at the step level.
As shown in Figure~\ref{fig:parallel_degree_timeseries_even}, when bursty arrivals create contention, \name dynamically raises the SP degree for computationally intensive, urgent requests to shorten their critical path and reduce SLO violation risk. 
Conversely, it scales down the degree for less urgent requests steps while maintain SLO Attainment Ratio. 
This fine-grained, adaptive parallelism is how \name handles burstiness and achieves superior efficiency and responsiveness compared to rigid, fixed-degree systems.


\begin{figure}[t]
    \centering
    \begin{subfigure}[b]{\columnwidth}
        \centering
        \includegraphics[width=\linewidth]{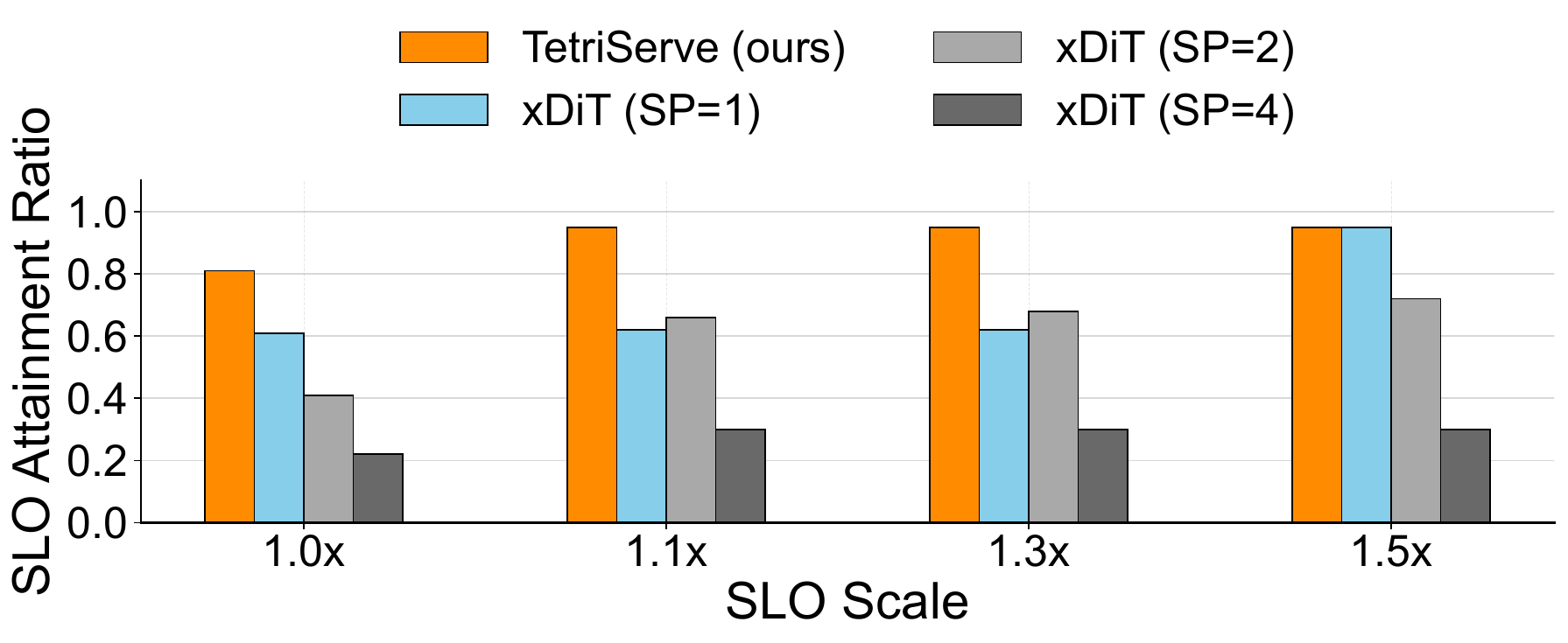}
        \caption{SAR vs. SLO Scale (SD3, Uniform mix)}
        \label{fig:sd3_goodput:even}
    \end{subfigure}
    \begin{subfigure}[b]{\columnwidth}
        \centering
        \includegraphics[width=\linewidth]{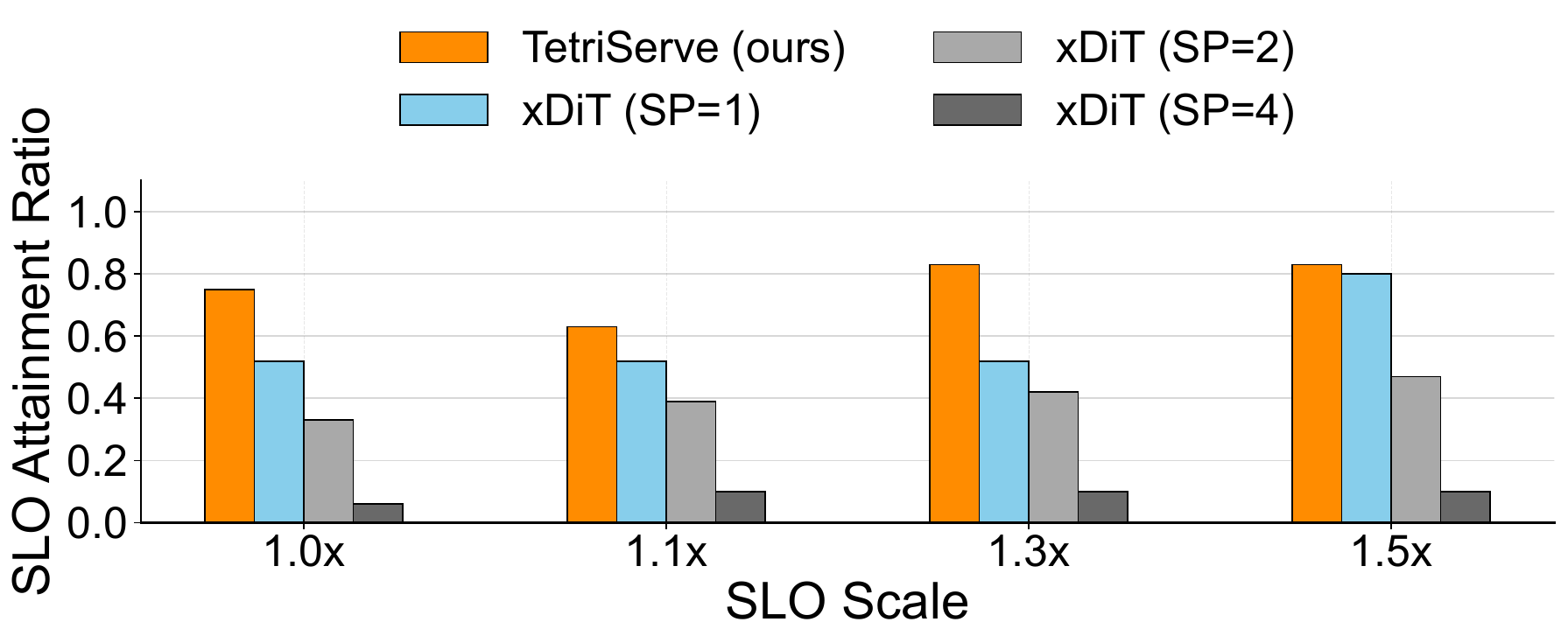}
        \caption{SAR vs. SLO Scale (SD3, Skewed mix)}
        \label{fig:sd3_goodput:large}
    \end{subfigure}
    \caption{\name's performance on the Stable Diffusion 3 (SD3) model. The plots show the SLO Attainment Ratio (SAR) as a function of SLO Scale for the Uniform mix (left) and Skewed mix (right) on 4$\times$A40 GPUs. In both workloads, \name consistently outperforms all xDiT variants}
    \label{fig:sd3_goodput}
\end{figure}

\subsection{Sensitivity Analysis}
\label{sec:eval:sensitivity}


\paragraph{Different GPU Settings and Models.}


On SD3, trends align with FLUX. In both the Uniform mix (Figure~\ref{fig:sd3_goodput}\subref{fig:sd3_goodput:even}) and Skewed mix (Figure~\ref{fig:sd3_goodput}\subref{fig:sd3_goodput:large}), \name achieves the highest SAR across all SLO scales, with the largest margins at tight SLOs (1.0$\times$). 
As SLOs loosen, fixed SP2 and SP4 improve but remain below \name, while fixed SP1 underutilize and plateau. 
This indicates the benefits generalize to a different DiT architecture. 
On the A40 cluster, NVLink links GPUs only in pairs; at SP=4, collectives traverse PCIe, and even at SP=2 poor placement can cross PCIe. For SD3 this communication path becomes the bottleneck, so SP2 and SP4 perform notably worse than on H100.


\begin{figure}[!t]
    \centering
    \includegraphics[width=0.98\columnwidth]{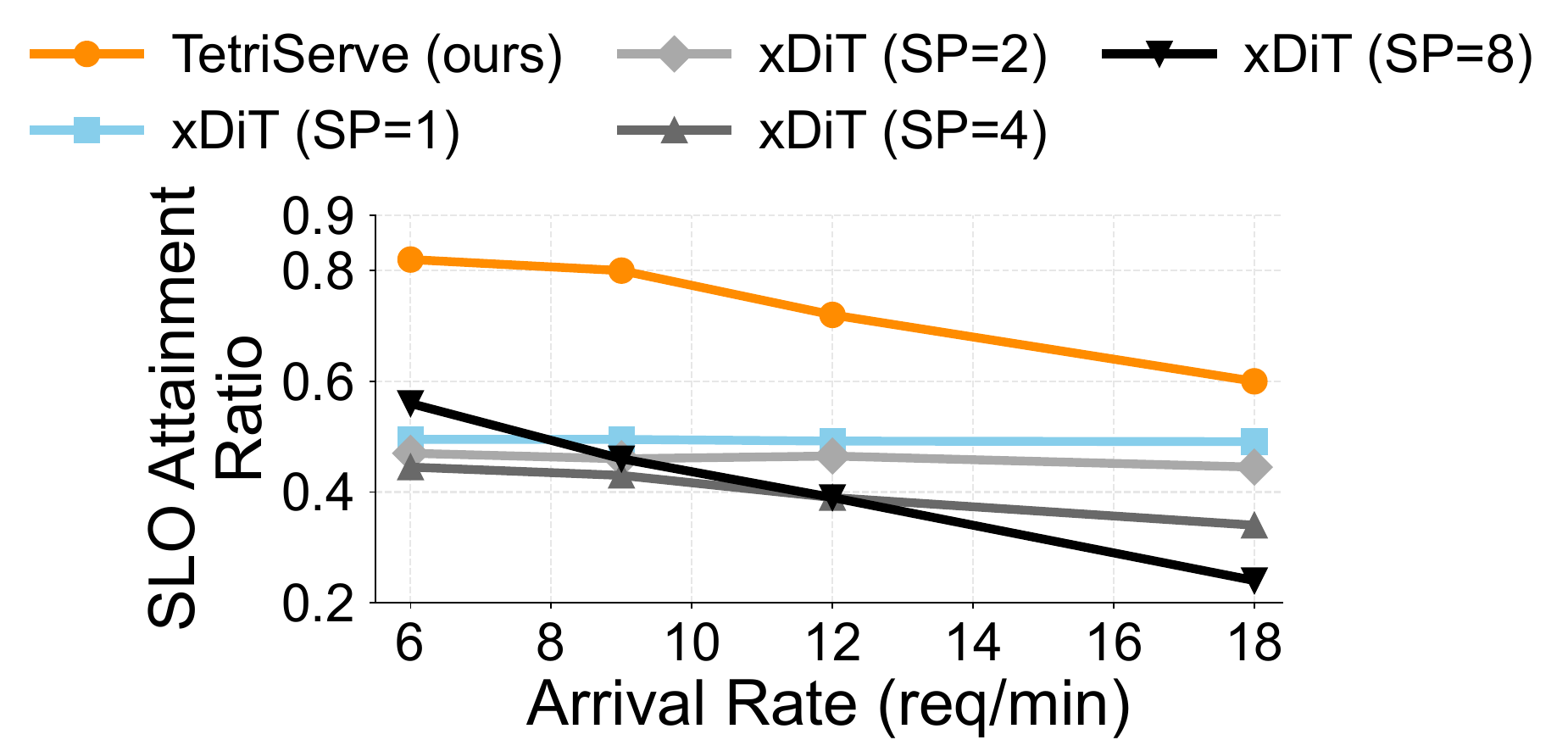}
    \caption{SLO Attainment Ratio vs. arrival rate under the Uniform mix (SLO Scale=1.0x). \name gracefully handles increasing load, maintaining a high SAR.}
    \label{fig:sar_vs_rate}
\end{figure}

\paragraph{Arrival Rate.}
Figure~\ref{fig:sar_vs_rate} shows the SAR of different scheduling strategies under the Uniform mix with a tight SLO of 1.0$\times$ as the arrival rate increases from 6 to 18 req/min. 
\name demonstrates superior performance across the full range of arrival rates. At low-to-medium rates, \name maintains a consistently high SAR, while fixed-parallelism strategies already show signs of degradation. 
At high arrival rates, where the system is under heavy load, \name's SAR remains relatively high, showcasing graceful degradation. 

\begin{figure}[!t]
    \centering
    \includegraphics[width=0.95\columnwidth]{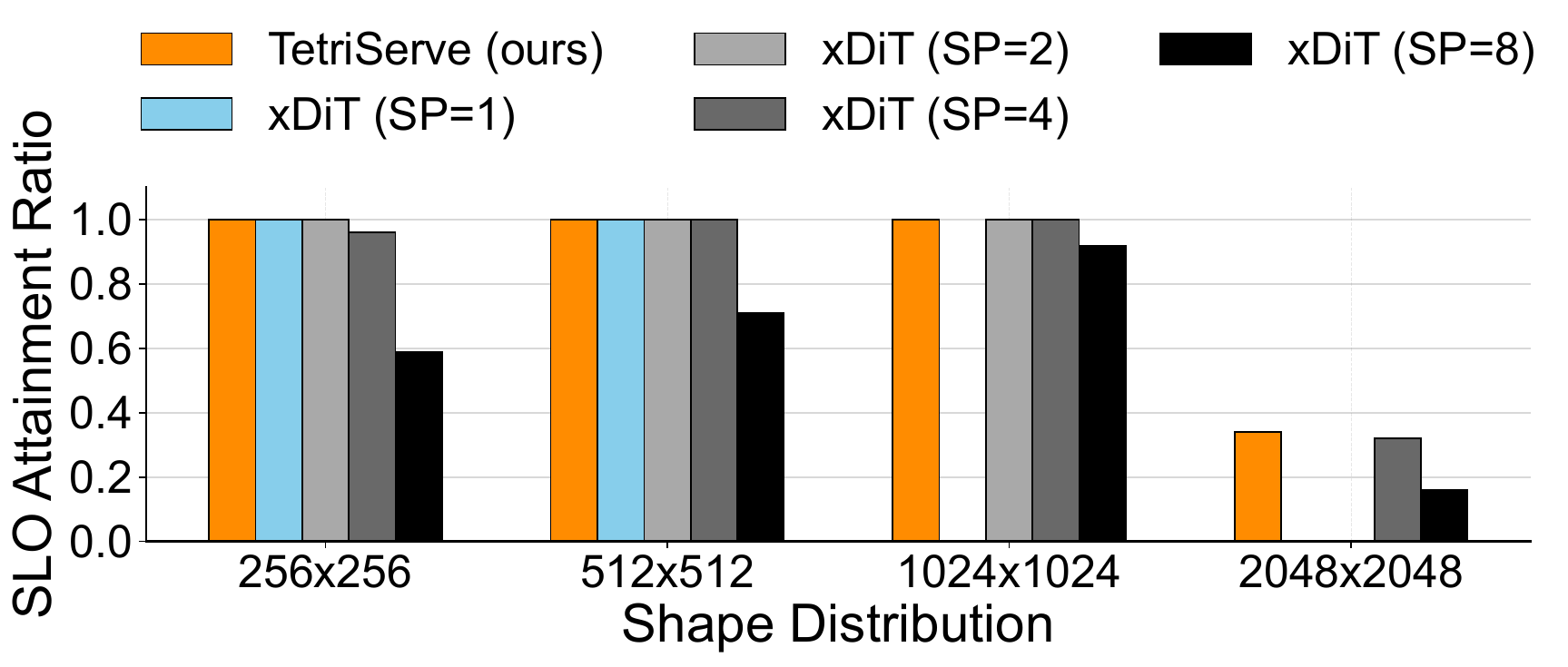}
    \caption{SLO Attainment Ratio for homogeneous workloads at 12 req/min with a 1.5x SLO Scale. Each group of bars represents a workload with only one resolution type. \name consistently achieves the highest SAR across all resolutions.}
    \label{fig:sar_homogeneous}
\end{figure}

\paragraph{Homogeneous Resolutions.}  
To isolate the effect of input resolution on parallelism strategies, we evaluate homogeneous workloads containing only a single resolution. Figure~\ref{fig:sar_homogeneous} shows the SLO Attainment Ratio (SAR) for workloads consisting of only one resolution type at an arrival rate of 12 req/min and an SLO Scale of 1.5x. Even in these simplified scenarios, \name still achieves the highest SAR across all resolution types. 
This demonstrates that \name's adaptive scheduling is effective not only for mixed workloads but also for homogeneous ones, as it can still optimize resource allocation to better meet deadlines.

\begin{figure}[!t]
    \centering
    \includegraphics[width=0.95\columnwidth]{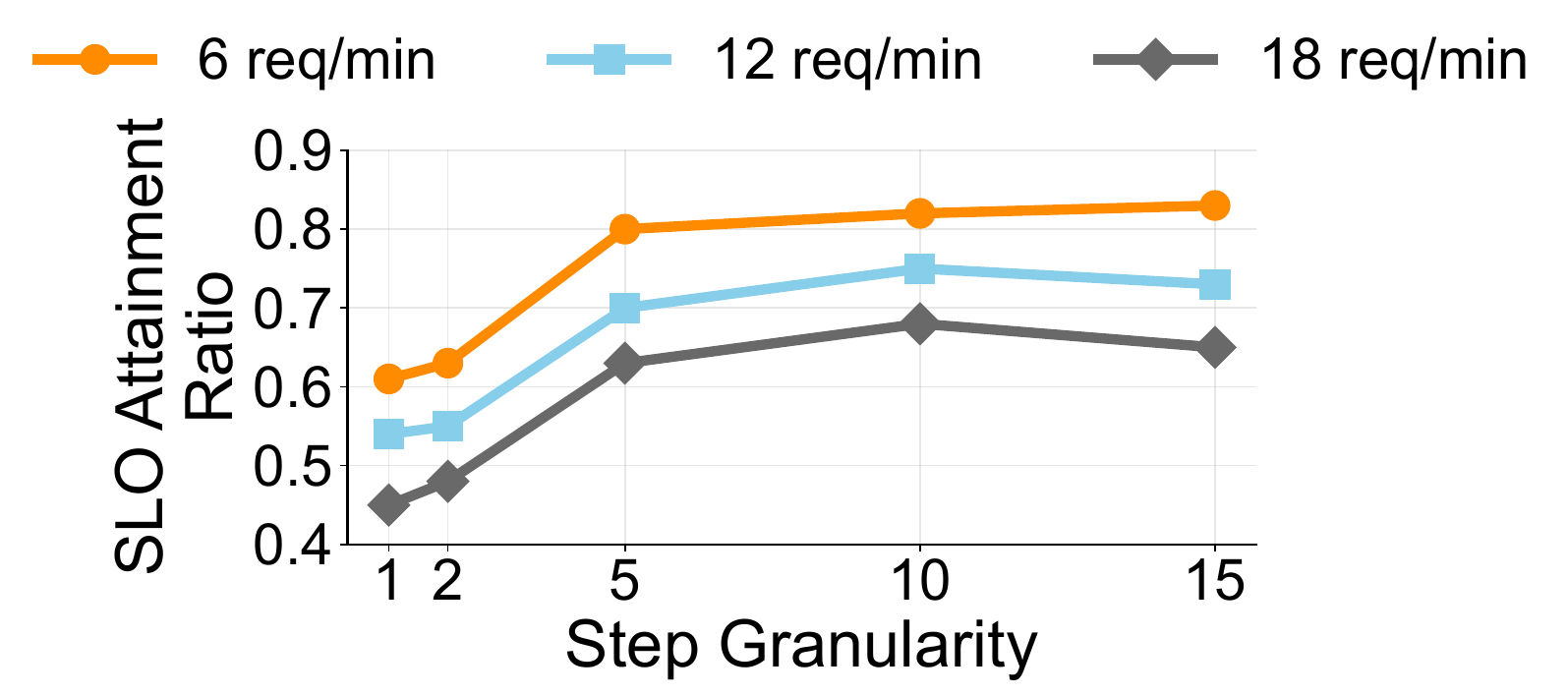}
    \caption{Sensitivity of SLO Attainment Ratio to step granularity and arrival rate under the Uniform mix (SLO Scale=1.0x). A moderate granularity (5/10 steps) provides the most robust performance as system load increases, balancing scheduling flexibility and overhead.}
    \label{fig:sensitivity_slo_meet_ratio_even_1x_sensitivity}
\end{figure}

\paragraph{Step Granularity.}
We examine the impact of step granularity, which defines how frequently \name can reschedule and change the degree of parallelism for an in-flight request. This presents a fundamental trade-off: fine-grained control (e.g., every 1-2 steps) offers maximum flexibility at the cost of high scheduling overhead, while coarse-grained control (e.g., every 10 steps) minimizes overhead but creates longer, non-preemptible execution blocks that reduce adaptability. Figure~\ref{fig:sensitivity_slo_meet_ratio_even_1x_sensitivity} illustrates this trade-off under the Uniform mix (SLO Scale=1.0x) across different arrival rates. At low rates, performance is less sensitive to granularity. However, as load increases, a moderate granularity of 5 steps proves most robust, balancing adaptability and overhead. Very fine-grained control (1 step) suffers from excessive overhead, while coarse-grained control (10 steps) is too inflexible to handle preemption, leading to lower SLO attainment.


\begin{table}[t]
    \centering
    \small 
    \caption{Latent transfer overhead as a percentage of inference step latency. Across all configurations, the overhead is negligible ($<0.05\%$).}
    \label{tab:reconfiguration_overhead_compact}
    \begin{tabular}{l|cccc}
        \toprule
        \textbf{Batch Size} & \textbf{256$\times$256} & \textbf{512$\times$512} & \textbf{1024$\times$1024} & \textbf{2048$\times$2048} \\
        \midrule
        BS = 1 & 0.03\% & 0.03\% & 0.04\% & 0.01\% \\
        BS = 2 & 0.04\% & 0.03\% & 0.05\% & 0.02\% \\
        BS = 4 & 0.04\% & 0.05\% & 0.03\% & 0.01\% \\
        \bottomrule
    \end{tabular}
\end{table}

\paragraph{Parallel Reconfiguration Overhead.}
\name performs step-level scheduling, which requires transferring intermediate latent representations and metadata across GPU groups when parallelism changes between steps. Table~\ref{tab:reconfiguration_overhead_compact} quantifies this parallel reconfiguration overhead as a percentage of per step inference latency across varying resolutions and batch sizes. We observe that the overhead is consistently negligible, accounting for at most 0.05\% of step latency in all configurations. As a result, \name's scheduler can safely ignore latent transfer time in deadline accounting without affecting SLO accuracy.


\begin{table}[t]
\centering
\small
\caption{\textbf{Ablation of scheduling mechanisms.}
GPU Placement Preservation reduces inter-round stalls by keeping requests on the same GPU set; Elastic Scale-up opportunistically reallocates idle GPUs to requests that benefit from extra parallelism.}
\label{tab:ablation_study}
\setlength{\tabcolsep}{4pt}
\begin{subtable}[t]{\linewidth}
  \centering
  \caption{\textbf{Uniform Mix}.}
  \begin{tabular}{lcc}
    \toprule
    Variant & SLO = 1.0$\times$ & SLO = 1.5$\times$ \\
    & SAR $\uparrow$ / Mean Lat. $\downarrow$ & SAR $\uparrow$ / Mean Lat. $\downarrow$ \\
    \midrule
    \name schedule              & 0.54 / 4.45 & 0.74 / \textbf{4.81} \\
    + Placement              & 0.56 / 3.96 & 0.69 / 5.14 \\
    + Elastic Scale-Up        & \textbf{0.63} / \textbf{3.89} & \textbf{0.78} / 4.83 \\
    \bottomrule
  \end{tabular}
\end{subtable}

\vspace{0.4em}

\begin{subtable}[t]{\linewidth}
  \centering
  \caption{\textbf{Skewed Mix}.}
  \begin{tabular}{lcc}
    \toprule
    Variant & SLO = 1.0$\times$ & SLO = 1.5$\times$ \\
    & SAR $\uparrow$ / Mean Lat. $\downarrow$ & SAR $\uparrow$ / Mean Lat. $\downarrow$ \\
    \midrule
    \name schedule              & 0.27 / 8.43 & 0.38 / 9.92 \\
    + Placement              & 0.31 / \textbf{7.64} & 0.45 / 8.16 \\
    + Elastic Scale-Up        & \textbf{0.36} / 7.68 & \textbf{0.55} / \textbf{7.71} \\
    \bottomrule
  \end{tabular}
\end{subtable}
\end{table}

\subsection{Ablation Study}
\label{sec:eval:ablation_study}

\name includes two practical mechanisms on top of the round-based DP scheduler: (i) \emph{GPU Placement Preservation}, which keeps a request on the same GPU set across rounds whenever possible to avoid remapping stalls; and (ii) \emph{Elastic Scale-up}, which makes use of idle GPUs after placement and temporarily grants extra GPUs to requests that benefit from higher parallelism.
To quantify their impact, we ablate these components under two SLO scales (1.0$\times$ and 1.5$\times$) on two workload mixes: Uniform and Skewed. Table~\ref{tab:ablation_study} reports the SLO Attainment Ratio and mean latency.

Overall, both mechanisms are important for improving serving efficiency.
GPU Placement Preservation improves SAR and/or mean latency in most settings by avoiding remapping overhead and enabling immediate progress at round boundaries, while Elastic Scale-up consistently increases SAR (up to +0.11 absolute on Skewed mix at 1.5$\times$) and typically further reduces mean latency by utilizing idle GPUs.
Consequently, enabling both GPU placement preservation and Elastic Scale-up achieves the best SLO Attainment Ratio across all tested scenarios, while also improving latency compared to disabling these optimizations.
\section{Related Work}\label{sec:related}

\paragraph{LLM Serving Frameworks.}
LLM serving systems \cite{PagedAttention-sosp23,SGLang-neurips24} are not directly applicable to DiT workloads. 
LoongServe~\cite{loongserve-sosp24} optimizes prefill-decode stages for long-context LLMs, while PrefillOnly~\cite{prefillonly-sosp25} targets memory efficiency for short, prefill-intensive requests. 
Neither suits the multi-step, stateless inference pattern of DiTs.

\paragraph{DiT Inference and Serving.}
DiT-specific serving systems are still emerging. 
xDiT~\cite{xDiT-arxiv24} uses fixed sequence parallelism, which is inefficient for heterogeneous workloads. 
DDiT~\cite{ddit-arxiv25} targets video generation and maximizes throughput rather than meeting SLOs. 
\name uniquely prioritizes SLO attainment for heterogeneous requeststhrough cost-model-driven scheduling.

\paragraph{Text-to-Image Caching.}
Several systems accelerate text-to-image diffusion via caching.
AsyncDiff~\cite{asyncdiff-neurips24} parallelizes diffusion through asynchronous denoising cross requests.
Caching-based approaches exploit reuse across prompts or adapters, including approximate latent caching in Nirvana~\cite{ApproximateCaching-nsdi24}, layer-level caching~\cite{learningtocache-neurips24}, final image caching~\cite{modm-asplos26}, workflow-aware reuse~\cite{katz-atc25}, and patch-level reuse~\cite{mixfusion-ppopp26}.
These techniques reduce redundant computation; \name addresses an orthogonal dimension by scheduling GPU parallelism across concurrent requests and could integrate these methods for further gains.

\paragraph{Resource Scheduling.}
In VM allocation frameworks~\cite{VMAllocation-mlsys23}, machine count is fixed at admission. 
GPU schedulers like Gavel~\cite{narayanan2020heterogeneous}, Tiresias~\cite{tiresias:nsdi19}, and AlloX~\cite{allox:eurosys20} focus on job placement and fairness but require users to specify parallelism.
In contrast, \name treats parallelism as a scheduling decision, dynamically adjusting GPU degree at step granularity based on deadlines and scaling efficiency.

\section{Conclusion}\label{sec:conclusion}

We presented \name, a deadline-aware round-based DiT serving system that addresses the challenge of meeting SLOs under heterogeneous workloads. 
\name dynamically adapts parallelism at the \textit{step level}, guided by a profiling-driven cost model and a deadline-aware scheduling algorithm. 
Extensive evaluation shows that \name consistently outperforms fixed-parallelism baselines, achieving up to 32\% higher SLO attainment and robust performance across varying resolutions, workload distributions, and arrival rates.

\label{EndOfPaper}

\section*{Acknowledgements}
We thank the ASPLOS reviewers, as well as members of SymbioticLab and UseSysLab, for their helpful feedback.
This work was supported in part by NSF grants CCF-2450085, CNS-2106184, CNS-2214272 and CNS-2106751, and by grants from Ford and Cisco.


\bibliographystyle{ACM-Reference-Format}
\bibliography{ref}

\appendix





\clearpage

\section{NP-Hardness of DiT Serving}
\label{appendix:appendix-proofs}
We prove NP-hardness for the DiT serving problem defined in \name,
which maximizes the number of requests that complete by
deadlines under GPU capacity constraints. 

Let us first define the decision problem \textsc{DiT-Serving-Decision}:
given an instance, and an integer target $B$, decide whether there exists a schedule in which at least $B$ requests meet their deadlines. 
This is the natural decision version of \name's objective $\max \sum_i I_i$. 

Bar-Noy et al. \cite{bar1999approximating, garey1977two} state that the following real-time (RT) scheduling feasibility decision problem (\textsc{RT-Feasibility}) is
NP-hard in the strong sense: on a \emph{single} machine, given jobs with release
times $r_i$, deadlines $d_i$, and processing times $l_i$, decide whether \emph{all} jobs can be scheduled
within their time windows. 
Since \textsc{RT-Feasibility} is strongly NP-hard, it remains NP-hard even when all numeric parameters are bounded by a polynomial in the input size. 
Therefore, $T_{\max}=\max_i d_i$ is polynomially bounded, and our time-indexed reduction is polynomial-time.

\paragraph{Reduction to DiT serving with $\mathcal{K}=\{1\}$.}
Given a \textsc{RT-Feasibility} instance \cite{bar1999approximating} with jobs $i=1,\dots,n$ and parameters $(r_i,d_i,l_i)$, let us construct a single-step DiT instance as follows: \(N := 1, R := n, S_i := 1, K := \{1\}, \texttt{arrival\_time}(i) := r_i, D_i := d_i, T_i(1) := l_i.\) 
Set the throughput target $B:=n$. 

Equivalently, in \name's single-step time-indexed formulation with variables $x_{i,t,k}$ and constraints (1)--(5), we restrict to $k=1$ and $N=1$, and disallow infeasible start times by setting $x_{i,t,1}=0$ whenever $t<r_i$ or $t+l_i>d_i$.

\paragraph{Correctness.}
(\(\Rightarrow\)) If the \textsc{RT-Feasibility} instance is feasible, let $s_i$
be the start time of job $i$ in a feasible single-machine schedule. 
Schedule each corresponding DiT request $i$ to start at time $s_i$ using one GPU. 
All requests meet deadlines, so $\sum_i I_i = n \ge B$.

(\(\Leftarrow\)) If the constructed DiT instance has a schedule with
$\sum_i I_i \ge n$, then all $n$ requests meet deadlines. 
Since $N=1$ and each request uses one GPU, the capacity constraint implies no two requests overlap.
Thus the chosen start times form a feasible non-preemptive single-machine schedule for all jobs in the original \textsc{RT-Feasibility} instance.

Therefore, we can convert any \textsc{RT-Feasibility} instance into a
\textsc{DiT-Serving-Decision} instance in polynomial time such that a feasible schedule exists in the former iff one exists in the latter.
\textsc{DiT-Serving-Decision} is NP-hard even for the restricted case $S_i=1$ and $\mathcal K=\{1\}$; consequently, the general multi-step DiT serving problem is NP-hard.

\begin{table}[t]
    \centering
    \small
    \setlength{\tabcolsep}{8pt}
    \renewcommand{\arraystretch}{1.15}

    \begin{subtable}[t]{0.47\linewidth}
        \centering
        \label{tab:sched-overhead-4gpus}
        \begin{tabular}{c c}
            \toprule
            \textbf{\# Reqs} & \textbf{Time (s)} \\
            \midrule
            1  & \textit{<0.01} \\
            2  & \textit{0.27} \\
            3  & \textit{52.56} \\
            4  & \textit{>60.00} \\
            \bottomrule
        \end{tabular}
        \caption{4 GPUs}
    \end{subtable}
    \hfill
    \begin{subtable}[t]{0.47\linewidth}
        \centering
        \label{tab:sched-overhead-8gpus}
        \begin{tabular}{c c}
            \toprule
            \textbf{\# Reqs} & \textbf{Time (s)} \\
            \midrule
            1   & \textit{0.02} \\
            2   & \textit{11.12} \\
            3   & \textit{>60.00} \\
            4   & \textit{>60.00} \\
            \bottomrule
        \end{tabular}
        \caption{8 GPUs}
    \end{subtable}

    \caption{\textbf{Scheduling overhead of exhaustive search.} Control plane scheduling time under different GPU budgets and queue sizes. \name remains lightweight: it takes \textbf{\emph{<0.01\,s}} compared to exhaustive search following the same settings, enabling online scheduling in practice.}
    \label{tab:scheduling_overhead}
\end{table}

\section{Scheduling Overhead Analysis}
\label{appendix:scheduling-overhead}

To validate the necessity of \name's heuristic approach, we quantify the computational cost of finding a globally optimal schedule via exhaustive search. As established in Appendix~\ref{appendix:appendix-proofs}, the underlying step-level scheduling problem is NP-hard.

\textbf{Experimental Setup.} We implement an exact baseline solver that enumerates the complete decision space to maximize SLO attainment. The solver explores two dimensions of complexity for each request: (1) all feasible sequence-parallel degrees per diffusion step (e.g., $k\in\{1,2,4,8\}$), and (2) all valid permutations of physical GPU mapping for those degrees. The objective is to identify the schedule with the highest SLO attainment, using minimum total GPU hours as a tie-breaker. We measure the wall clock latency required to generate a single scheduling plan using an AMD EPYC 7513 32-Core CPU, varying the queue depth ($R$) under fixed GPU budgets of $N \in \{4, 8\}$.

\textbf{Results.} Table~\ref{tab:scheduling_overhead} presents the scheduling overhead. The baseline exhibits immediate combinatorial explosion: with a budget of 8 GPUs, optimally scheduling merely three requests exceeds a 60-second timeout. This intractability stems from the factorial growth of permutation possibilities as the number of available GPUs increases. In contrast, \name maintains a decision latency of \emph{<10\,ms}. These results confirm that exhaustive optimization is prohibitive for online serving, necessitating the efficient round-based planning strategy employed by \name.

\end{document}